\documentclass{article}

\usepackage[T1]{fontenc}
\usepackage{lmodern,fullpage}

\usepackage{parskip}

\newcommand{\vectfield}{F}
\newcommand{\vectfieldidx}[1]{\vectfield_{#1}}
\newcommand{\partder}[1]{\vectfield_{d,#1}}
\newcommand{\WW}{W}
\newcommand{\VV}{V}
\newcommand{\ww}{\boldsymbol{w}}
\newcommand{\wwi}[1]{\boldsymbol{w}_{#1}}
\newcommand{\vvi}[1]{\boldsymbol{v}_{#1}}
\newcommand{\uui}[1]{\boldsymbol{u}_{#1}}
\newcommand{\GGG}{\Gamma}
\newcommand{\bxi}{\boldsymbol{\xi}}

\usepackage[utf8]{inputenc} 
\usepackage[T1]{fontenc}    
\usepackage{hyperref}       
\usepackage{url}            
\usepackage{booktabs}       
\usepackage{amsfonts}       
\usepackage{nicefrac}       
\usepackage{microtype}      
\usepackage{xcolor}         
\usepackage{breqn}
\usepackage{lscape}
\usepackage{setspace}
\usepackage{microtype,bm}
\usepackage{xspace}
\usepackage{algorithm}
\usepackage{mathtools}
\usepackage{amssymb}
\usepackage{xpatch}
\usepackage{enumitem}

\usepackage[nointegrals]{wasysym}

\newtheorem{definition}{Definition}

\newtheorem{rem}{Remark}
\newtheorem{example}{Example}

\newtheorem{theorem}{Theorem}

\newtheorem{proposition}{Proposition}
\newtheorem{lemma}{Lemma}

\newcommand{\base}{b}

\xpatchcmd{\proof}{\itshape}{\normalfont\proofnameformat}{}{}
\newcommand{\proofnameformat}{\bfseries}

\newcommand{\real}{\mathbb{R}}

\usepackage{prettyref}
\newcommand{\pref}[1]{\prettyref{#1}}

\newcommand{\savehyperref}[2]{\texorpdfstring{\hyperref[#1]{#2}}{#2}}
\newrefformat{eq}{\savehyperref{#1}{\textup{(\ref*{#1})}}}
\newrefformat{eqn}{\savehyperref{#1}{Equation~\ref*{#1}}}
\newrefformat{lem}{\savehyperref{#1}{Lemma~\ref*{#1}}}
\newrefformat{obs}{\savehyperref{#1}{Observation~\ref*{#1}}}
\newrefformat{def}{\savehyperref{#1}{Definition~\ref*{#1}}}
\newrefformat{line}{\savehyperref{#1}{line~\ref*{#1}}}
\newrefformat{thm}{\savehyperref{#1}{Theorem~\ref*{#1}}}
\newrefformat{corr}{\savehyperref{#1}{Corollary~\ref*{#1}}}
\newrefformat{sec}{\savehyperref{#1}{Section~\ref*{#1}}}
\newrefformat{exam}{\savehyperref{#1}{Example~\ref*{#1}}}
\newrefformat{prop}{\savehyperref{#1}{Proposition~\ref*{#1}}}
\newrefformat{table}{\savehyperref{#1}{Table~\ref*{#1}}}
\newrefformat{app}{\savehyperref{#1}{Appendix~\ref*{#1}}}
\newrefformat{ass}{\savehyperref{#1}{Assumption~\ref*{#1}}}
\newrefformat{tbl}{\savehyperref{#1}{Table~\ref*{#1}}}
\newrefformat{ex}{\savehyperref{#1}{Example~\ref*{#1}}}
\newrefformat{fig}{\savehyperref{#1}{Figure~\ref*{#1}}}
\newrefformat{alg}{\savehyperref{#1}{Algorithm~\ref*{#1}}}
\newrefformat{rem}{\savehyperref{#1}{Remark~\ref*{#1}}}
\newrefformat{conj}{\savehyperref{#1}{Conjecture~\ref*{#1}}}
\newrefformat{prop}{\savehyperref{#1}{Proposition~\ref*{#1}}}
\newrefformat{proto}{\savehyperref{#1}{Protocol~\ref*{#1}}}
\newrefformat{prob}{\savehyperref{#1}{Problem~\ref*{#1}}}
\newrefformat{claim}{\savehyperref{#1}{Claim~\ref*{#1}}}

\newcommand{\ploss}{{\cL}} 

\newcommand{\is}[1]{[#1]}

\newcommand{\ibr}[1]{{[#1]}}

\newcommand{\fx}{\mathfrak{x}}
\newcommand{\fy}{\mathfrak{y}}

\newcommand{\qed}{$\square$}
\newcommand{\grad}[1]{{\nabla {#1}}}
\newcommand{\balpha}{\boldsymbol{\alpha}}
\newcommand{\bbeta}{\boldsymbol{\beta}}
\newcommand{\defoo}{~\dot{=}~}
\newcommand{\fpscoeff}[2]{c_{#1,#2}}
	
\def\ddefloop#1{\ifx\ddefloop#1\else\ddef{#1}\expandafter\ddefloop\fi}
\def\ddef#1{\expandafter\def\csname 
	bb#1\endcsname{\ensuremath{\mathbb{#1}}}}
\ddefloop ABCDEFGHIJKLMNOPQRSTUVWXYZ\ddefloop
\def\ddef#1{\expandafter\def\csname 
	#1\endcsname{\ensuremath{{\bm{#1}}}}}
\ddefloop ABCDEFGHIJKLMNOPQRSTUVWXYZ\ddefloop
\def\ddef#1{\expandafter\def\csname 
	#1\endcsname{\ensuremath{{\bm{#1}}}}}
\ddefloop abcdefghijklmnopqrstuvwxyz\ddefloop
\def\ddefloop#1{\ifx\ddefloop#1\else\ddef{#1}\expandafter\ddefloop\fi}
\def\ddef#1{\expandafter\def\csname 
	b#1\endcsname{\ensuremath{\mathbf{#1}}}}
\ddefloop ABCDEFGHIJKLMNOPQRSTUVWXYZ\ddefloop
\def\ddef#1{\expandafter\def\csname 
	c#1\endcsname{\ensuremath{\mathcal{#1}}}}
\ddefloop ABCDEFGHIJKLMNOPQRSTUVWXYZ\ddefloop
\def\ddef#1{\expandafter\def\csname 
	f#1\endcsname{\ensuremath{\mathfrak{#1}}}}
\ddefloop ABCDEFGHIJKLMNOPQRSTUVWXYZ\ddefloop
\def\ddef#1{\expandafter\def\csname 
	h#1\endcsname{\ensuremath{\widehat{#1}}}}
\ddefloop ABCDEFGHIJKLMNOPQRSTUVWXYZ\ddefloop
\def\ddef#1{\expandafter\def\csname 
	hc#1\endcsname{\ensuremath{\widehat{\mathcal{#1}}}}}
\ddefloop ABCDEFGHIJKLMNOPQRSTUVWXYZ\ddefloop
\def\ddef#1{\expandafter\def\csname 
	t#1\endcsname{\ensuremath{\widetilde{#1}}}}
\ddefloop ABCDEFGHIJKLMNOPQRSTUVWXYZ\ddefloop
\def\ddef#1{\expandafter\def\csname 
	tc#1\endcsname{\ensuremath{\widetilde{\mathcal{#1}}}}}
\ddefloop ABCDEFGHIJKLMNOPQRSTUVWXYZ\ddefloop

\newcommand{\inner}[1]{\langle #1 \rangle}

\newcommand{\RR}{\mathbb{R}}
\newcommand{\EE}{\mathbb{E}}

\newcommand{\bones}{\bm{1}}
\newcommand{\pint}{\mbox{$\mathbb{N}$}}

\DeclarePairedDelimiter{\abs}{\lvert}{\rvert} %
\DeclarePairedDelimiter{\brk}{[}{]}
\DeclarePairedDelimiter{\crl}{\{}{\}}
\DeclarePairedDelimiter{\prn}{(}{)}
\DeclarePairedDelimiter{\nrm}{\|}{\|}
\newcommand{\defeq}{\coloneqq}
\newcommand{\dd}{\mbox{$\;|\;$}}
\newcommand{\arr}{{\rightarrow}}
\newcommand{\btheta}{\boldsymbol{\theta}}
\newcommand{\sgn}{\mbox{sgn}}

\title{Annihilation of Spurious Minima \\in Two-Layer ReLU Networks}

\author{%
	{Yossi Arjevani}\\
	The Hebrew University\\
	\texttt{yossi.arjevani@gmail.com}
	\and
	Michael Field\\
	UC Santa Barbara\\
	\texttt{Mike.field@gmail.com}
}

\date{}

\begin{document}
	\newcommand{\bigfamily}[3]{\mathfrak{C}^{\mathrm{#1}}_{#2,#3}}
	\newcommand{\smallfamily}[3]{\mathfrak{c}^{\mathrm{#1}}_{#2,#3}}
	\newcommand{\txi}{\xi}
	\newcommand{\tbxi}{\boldsymbol{\xi}}
	\newcommand{\tXi}{\Xi}
	
	\maketitle
	\begin{abstract}
		We study the optimization problem associated with fitting two-layer ReLU
		neural networks with respect to the squared loss, where labels are generated 
		by a target network. Use is made of the rich symmetry structure to develop a
		novel set of tools for studying the mechanism by which over-parameterization
		annihilates spurious minima. Sharp analytic estimates  are obtained
		for the loss and the Hessian spectrum at different minima, and it is proved 
		that adding neurons can turn symmetric spurious minima into
		saddles; minima of lesser symmetry require more neurons. Using Cauchy's
		interlacing theorem, we prove the existence of descent directions
		in certain subspaces arising from the symmetry structure of the loss
		function. This analytic approach uses techniques, new to the field, from 
		algebraic geometry, 	representation theory and symmetry breaking, and 
		confirms rigorously the
		effectiveness of over-parameterization in making the associated loss
		landscape accessible to gradient-based methods. For a fixed number of 
		neurons and inputs, the spectral results remain true under symmetry
		breaking perturbation of the target.
	\end{abstract}


\section{Introduction}
An outstanding question in deep learning (DL) concerns the ability of
simple gradient-based methods to successfully train neural networks despite 
the
nonconvexity of the associated optimization problems. Indeed, nonconvex
optimization landscapes may have spurious (i.e., non-global local) minima 
with large
basins of attraction and this can cause a complete failure of these
methods. Evidence suggests that this problem can be
circumvented by the use of a large number of parameters in DL
models.
In view of the complexity exhibited by contemporary neural networks
and the absence of suitable analytic tools, much recent research has
focused on two-layer ReLU networks as a realistic starting point for a
theoretical study
\cite{brutzkus2017globally,chizat2018global,soltanolkotabi2018theoretical,
	mei2018mean,goldt2019generalisation,tian2020student,safran2021effects}.
The two-layer networks considered were typically of the form:
\begin{align} \label{net}
	f(\x; \WW, \balpha) \defeq
	\balpha^\top\!\varphi\prn{W\x},\quad
	~\WW\in 
	M(k,d),~\balpha \in \RR^k,
\end{align}
where $\varphi(z)\defeq\max\crl{0,z}$ is the ReLU function acting entrywise and 
$M(k,d)$ denotes the space of $k\times d$ matrices. In order to isolate the 
study of optimization-related obstructions due to
nonconvexity from issues pertaining to the expressive power of two-layer
networks, data has been often assumed to be fully realizable.
This was further motivated by hardness results which 
indicated a strict barrier inherent to the explanatory power of 
distribution-free approaches operating in complete generality 
\cite{blum1989training,brutzkus2017sgd,shamir2018distribution}. 
For the
squared loss, the resulting, highly nonconvex, expected loss~is
\begin{align}\label{opt:problem}
	\cL(\WW, \balpha) \defeq \frac{1}{2}\EE_{\x\sim 
		\cD}\brk*{\Big(f(\x; \WW, \balpha) -      f(\x; \VV, 
		\boldsymbol{\beta})\Big)^2},
\end{align}
where $\cD$ denotes a probability distribution over the input space, $\WW\in 
M(k,d)$, $\balpha \in \RR^k$ are the optimization variables, and $\VV\in 
M(d,d)$, $\bbeta \in \RR^d$ are fixed parameters.

The choice of the $d$-variate normal Gaussian distribution for the input
distribution has drawn a considerable interest, e.g.,
\cite{du2017gradient,zhong2017recovery,li2017convergence,tian2017analytical,
	ge2017learning,aubin2019committee,akiyama2021learnability}.
Empirically, it has been observed that as the number of neurons $k$ 
increases, the loss of models obtained using stochastic gradient 
descent (SGD) under Xavier initialization 
\cite{glorot2010understanding} decreases (see 
\cite{brutzkus2017sgd,safran2017spurious}).
The objective of this work is to study the mathematical mechanism behind 
this fundamental phenomenon. Using 
ideas based on \emph{symmetry breaking} (see \pref{sec:sym_break}), we are
able to employ techniques from representation theory and algebraic 
geometry to study how the loss landscape is transformed when the 
number of neurons is increased. A~key feature of our approach is the 
use of \emph{Puiseux series} in $d$ and path based techniques. For example, we 
may define the spectrum of the Hessian as Puiseux series for 
\emph{real} values of $d$ and so determine the values of $d$ 
(typically not integers) where eigenvalues change sign. These 
methods allow the proof of powerful analytic results: in this paper, 
we derive sharp analytic estimates of the loss and the Hessian
spectrum at local minima allowing us to analyze the mechanism 
whereby spurious minima are annihilated when the number of neurons 
$k$ is increased. Our contributions can be summarized~by
\begin{theorem} (Informal) \label{thm:main0}
	We describe several infinite families of spurious minima, partly 
	characterized by
	their symmetry, and prove that:\\
	$\bullet$ Adding rather few neurons can transform symmetric spurious 
	minima into
	saddles.\\  
	$\bullet$ The Hessian spectrum remains extremely skewed with $\Theta(d)$
	eigenvalues growing linearly with $d$, and $\Theta(kd)$ eigenvalues
	being $\Theta(1)$. \\
	$\bullet$ Increasing the number of neurons adds $\Theta(d)$ descent
	directions in a $\Theta(d)$-dimensional subspace dictated by the 
	symmetry structure (i.e., the isotypic decomposition) of the loss 
	function.
	\\
	$\bullet$ The loss remains (essentially) unchanged.
\end{theorem}
The formation of new decent directions allows gradient-based methods to 
escape spurious minima that exist in the original 
loss landscape ($k=d$), and so detect models of reduced loss. To indicate 
the subtly of the results given in \pref{thm:main0}, consider a family of 
spurious minima with symmetry $\Delta (S_{d-1}\times S_1)$ (the type II family 
described in \pref{sec:family}). Assume $k = d$. Spurious minima occur for 
$d \ge 6$. Adding one neuron results in these minima occurring for $d \ge 
8$ --- not 
promising. Adding two neurons annihilates \emph{all} these spurious minima 
and creates no new spurious minima. The mechanism is finite, cannot be 
inferred from a limiting case, and persists under forced symmetry 
breaking, and so applies beyond symmetric targets.
If instead we consider the unique family of 
spurious minima with isotropy $\Delta S_d$ (see \pref{sec:family}),
we find that adding just one neuron annihilates these spurious minima.  
This occurs through the appearance of a multiplicity $(d-1)$-eigenvalue 
associated to 
the standard representation of $S_d$ on $\real^{d-1}$ (all other 
eigenvalues remain positive). More precisely, if $k = d$ there are 3 
strictly positive Hessian eigenvalues associated to the 
standard representation of $S_d$, each with multiplicity $d-1$. Modulo 
$O(d^{-\frac{1}{2}})$ terms, they are $\frac{1}{4} - \frac{1}{2\pi}, 
\frac{1}{4}$, and $\frac{d}{4} + \frac{1}{4}$.
After the addition of one neuron ($k = d+1$), there are 4 Hessian 
eigenvalues associated to the
standard representation of $S_d$ and, modulo  $O(d^{-\frac{1}{2}})$ terms, 
these satisfy 
\[
\frac{d}{4} + \frac{1}{2}, \frac{1}{4}-\frac{1}{2\pi}, 
\frac{-1+\sqrt{5}}{4\pi}+\frac{1}{4} > 0 > 
\frac{-1-\sqrt{5}}{4\pi}+\frac{1}{4}.
\]
This example highlights the special role that the standard representation 
plays in the annihilation of spurious minima (see \pref{sec:over} and 
the concluding remarks). The sharp estimates of the Hessian spectrum further 
demonstrate how symmetry breaking enables a complete characterization of the 
dynamics of gradient-based methods, locally, in the vicinity of symmetric 
critical points. The dependence of 	such methods on stability of critical 
points therefore indicates that attempts for a global theory should be preceded 
by a good description of the mechanism by which spurious minima 
transform into saddles---the aim of this work. 

Next, we relate our results to the existing literature. 

\paragraph{Annihilation of spurious minima on account of over-parameterization.}
Existing methods for the analysis of optimization problem (\ref{opt:problem}) 
include: 
mean-field \cite{mei2018mean}, optimal transport \cite{chizat2018global}, 
NTK \cite{jacot2018neural,daniely2016toward,li2020learning} and the 
thermodynamic limit 	
\cite{goldt2019generalisation,aubin2019committee,goldt2019dynamics,oostwal2021hidden}.
These methods operate by passing to limiting regimes where the number of 
inputs or neurons is taken to infinity. A growing number of works has limited 
the explanatory power of such approaches  
\cite{yehudai2019power,ghorbani2020neural}. Approaches for addressing the 
loss landscapes in finite parameter regimes exist and include 
\cite{tian2020student} which obtains several generalities on critical points, 
and \cite{safran2021effects} which studies conditions under which a 
single neuron can be added in contrived way so as to turn a given minimum 
in $M(d,d)$ into a saddle in $M(d+1,d)$. However, as shown 
in the 	present work, families of minima in $M(d+1,d)$ exist. Indeed, 
families of minima of lesser symmetry are shown to require at least two 
additional neurons	before turning into	saddles.

\paragraph{Symmetry breaking in nonconvex loss landscapes.}
It has been recently found that the symmetry of spurious minima 
in optimization problem (\ref{opt:problem}) \emph{break the 
	symmetry} of global minima \cite{arjevanifield2019spurious} (see 
\pref{sec:sym_break} for a formal exposition). A similar 
phenomenon has been observed for tensor decomposition problems 
in \cite{arjevanifield2021tensor} and later studied in 
\cite{arjevanifield2022analytictensor}. The present work builds on 
methods developed in a line of work concerning this phenomenon of 
symmetry breaking. In \cite{ArjevaniField2020}, path based techniques are 
introduced to construct infinite families of critical points 
represented by Puiseux series in $d^{-1}$. In \cite{arjevanifield2020hessian}, 
results from the representation theory of the symmetric group are 
used, together with the Puiseux series result, to obtain precise 
analytic estimates on the Hessian spectrum. In~\cite{arjevanifield2021analytic}, it 
is shown that certain families of saddles 
transform into spurious minima at a fractional dimensionality. Moreover, the 
spectra of these families of spurious minima is shown to be identical to that 
of global minima to $O(d^{-\frac{1}{2}})$-order. In 
\cite{arjevanifield2022equivariant}, generic $S_d$-equivariant steady-state 
bifurcation is studied, emphasizing the complex geometry of the exterior square 
and the standard representations along which the minima studied in this 
work are created and annihilated.

The results above assume that the number of neurons is less or equal 
the number of inputs: $k \le d$. The present work concerns the 
over-parameterized case $k > d$ (additional terms used in related 
contexts are \emph{over-specified}, e.g., \cite{livni2014computational} and 
\emph{over-realized}, e.g., \cite{tian2020student}). The 
study of the technically more demanding case of 
over-parameterization 
requires new methods and ideas, which we describe~below.

\begin{itemize}[leftmargin=*]
	\item We develop a method which allows the expression of 
	eigenvalues in terms of the gradient entries. The approach is 
	used to evaluate eigenvalues of $\Theta(d^2)$-multiplicity analytically, 
	and reveals hidden algebraic relations between \emph{criticality} and 
	\emph{curvature}.

	\item Eigenvalues of $O(d)$-multiplicity are computed 
	numerically using estimates of the Puiseux series coefficients. The 
	estimates are obtained through numerical methods used either 
	directly for a system of equations corresponding to different 
	orders of the Puiseux series terms, or for a reduced system of 
	equations obtained by exploiting the geometric structure 
	of the problem.

	\item Lastly, Cauchy's interlacing theorem is used to reduce the complexity 
	of the computation of descent directions afforded by over-parameterization, 
	and yields a tight characterization of linear subspaces along which 
	spurious minima transform into saddles.
	
\end{itemize}
The methods are illustrated for eight families of critical points. 
Orthogonality of the target matrices is not required by the 	
symmetry-breaking framework; other choices of target matrices, 	
distributions, activation functions and	architectures have been considered 
in previous works, and are a topic of current research.

\paragraph{Organization of the paper.} The proof of \pref{thm:main0} is
in four parts: symmetry \& the loss function, families of symmetric minima, 
Hessian spectrum, and a change of stability under over-parameterization. 
Proofs 	and technical details are deferred to the appendix.

\section{Symmetry and the loss function} \label{sec:sym_break}  
The presentation of our results requires some familiarity with group
and representation theory. Key ideas and concepts are
introduced as needed. 

The \emph{symmetric group} $S_d$, $d\in\pint$, is the group of
permutations of $\ibr{d}\defoo \{1,\dots,d\}$. The \emph{orthogonal group}
$\text{O}(d)$ is the subgroup of all orthogonal linear maps on $\real^d$.
We may identify $S_d$ with the subgroup of $\text{O}(d)$ consisting of
permutation matrices.
Thus 
$S_d$ acts naturally on $[d]$ (as permutations) and orthogonally on 
$\real^d$ (as
permutation matrices).

For $k,d\in \pint$, there are natural actions of $S_k$ and $S_d$ on the 
space
$M(k,d)$ of $k \times d$ matrices: $S_k$ permutes rows, $S_d$ permutes 
columns. 
The loss function (\ref{opt:problem}) is invariant under row permutations: 
$\mathcal{L}(\sigma \WW)=\mathcal{L}(\WW)$, for all $\sigma \in 
S_{k}$---whatever the choice of $\VV,\boldsymbol{\beta}$. The product action 
of $S_k \times S_d$ on  $M(k,d)$ plays a central role in the study of
invariance properties of $\ploss$. If $A = [A_{ij}] \in M(k,d)$,
$(\pi,\rho)\in S_k \times S_d$, then
\begin{align}\label{eq: Gamma-action}
	(\pi,\rho)[A_{ij}] = [A_{\pi^{-1}(i),\rho^{-1}(j)}],\; \pi \in 
	S_k,\, 
	\rho \in S_d,\; (i,j) \in \ibr{k} \times \ibr{d}.
\end{align}
The action can be defined in terms of permutation matrices.	If $k\ge d$, 
the 
\emph{diagonal subgroup} $\Delta S_{d}$ of $S_k \times 
S_d$ is defined by $\Delta S_{d} = \{(g,g) \dd g \in S_d\subseteq S_k\}$. 
Clearly, $\Delta S_{d}\approx S_d$.

Henceforth assume $k \ge d$. Regard $M(d,d)$ as the linear subspace of 
$M(k,d)$
defined by appending $k-d$ zero rows to each matrix in $M(d,d)$.  Let $\VV\in 
M(k,d)$ denote the matrix determined by the identity matrix $I_d \in 
M(d,d)$.  By the orthogonal symmetry of the
Gaussian distribution, our results will hold for any $\VV$ determined by a 
matrix in $\text{O}(d)$. We shall not consider training processes but rather 
concrete instances of families of critical points, and so
throughout weights are assumed fixed. The weights of the second layer of 
critical points studied in the present work consists of positive weights 
and so, by the positive homogeneity of the ReLU activation, there is no loss of 
generality in assuming that the second layer of weights is set to ones, i.e., 
$\balpha=\bbeta=\mathcal{I}_{k,1}$, 
$\mathcal{I}_{i,j}$ being the $i\times j$-matrix with all entries 
equal to $1$.

Optimization problem (\ref{opt:problem}) 
has a rich symmetry	structure; for our choice of $\VV, \balpha$ and 
$\bbeta$, $\ploss$ is $S_k \times 
S_d$-invariant~\cite{arjevanifield2019spurious}. It is
natural to ask how the critical points of $\ploss$ reflect this
symmetry. Given $\WW \in M(k,d)$, the largest subgroup of
$S_k \times S_d$ fixing $\WW$ is called the \emph{isotropy} subgroup of $\WW$. 
The isotropy group quantifies 
the \emph{symmetry} of $\WW$.
If $k = d$, $\VV$ has isotropy group $\Delta S_d$ and every global minimizer 
of $\ploss$ lies on the $S_d \times S_d$-orbit of 
$V$~\cite{safran2021effects,ArjevaniField2020}. 
Empirically, if $k \ge d$, non-degenerate (i.e., no zero Hessian eigenvalues) 
spurious minima of $\ploss$ tend 
to be highly symmetric in 
that their isotropy groups are conjugate to large subgroups of $\Delta S_d$. 
Indeed, we suspect that for our choice of $\VV$, 
the isotropy of non-degenerate spurious minima is always non-trivial and 
conjugate to a subgroup of~$\Delta S_d$.

If $k > d$, the 
isotropy of $V$ is \emph{not} a subgroup of $\Delta S_d$--- it contains the 
subgroup $I_d \times S_{k-d}$ of row permutations.	Perhaps surprisingly, 
$\ploss$ is more regular at critical points of 
spurious minima than at points giving the global minimum: if $k = d$, 
analyticity of $\ploss$ at $\WW=\VV$~fails. If $k > d$, $V$ has a row of 
zeros and $\mathcal{L}$ is not	differentiable at $\WW = \VV$ 
\cite{brutzkus2017globally} (see also \cite{safran2021effects}). It may 
easily be shown that there is a 
$k-d$-dimensional compact connected $S_k \times S_d$-invariant simplicial 
complex $\Lambda\subset M(k,d)$ consisting of all matrices that define the 
global minimum value of zero (see \pref{sec:fossil}). Necessarily,  
$\Lambda$ contains the $S_k \times S_d$ orbit of $\VV$. At boundary points of 
$\Lambda$, $\mathcal{L}$ is not differentiable. The Hessian, defined on the 
interior of the simplex, is \emph{always} singular. Here our focus will 
always be on families of spurious minima with non-degenerate critical 
points.

%
%
%
%
%
%
%
%
%

\section{Families of minima: structure and basic properties} 
\label{sec:family}
\newcommand{\defoX}{\stackrel{\mathrm{def}}{=}}
Families of spurious minima often have characteristic properties. 
For 
example, the asymptotics in $d$ of the loss or their
chance of being detected by SGD. For a systematic study of the
optimization landscape of $\cL$, we need to categorize minima and
understand their distinctive analytic properties. Both isotropy and the 
notion of a \emph{regular family} play
an important role.  Throughout we assume $k= d+m$, for all $d \ge d_0$, 
where $m \ge 0$ is an integer~constant. 

If $G$ is a subgroup of $\Delta S_d$, let $M(k,d)^G = \{\WW \in M(k,d) \dd g\WW 
= \WW,\, \forall g \in G\}$
denote the \emph{fixed point space} for the action of $G$ on $M(k,d)$.
Every $S_d \times S_k$-equivariant vector field on $M(k,d)$ is tangent 
to $M(k,d)^G$ and so $\grad{\ploss}$ is tangent to~$M(k,d)^G$. Hence 
$\mathfrak{c} \in  M(k,d)^G$ is a critical point of 
$\ploss|M(k,d)^G$ iff $\mathfrak{c}$ is a critical point of 
$\ploss$. 
The inclusion $i_d: [d] \arr [d+1]$ induces a natural inclusion $i_d: S_d 
\subset S_{d+1}$, where $i_d(S_d)$ fixes $d+1$. More generally, given a 
positive integer $p$ and $d_0 > p$, we have a 
sequence of inclusions $i_{d,p} :S_{d-p} \times S_p \arr S_{d+1-p} \times 
S_p$, $d \ge d_0$.  
Identifying $S_d$ with $\Delta S_d$, a 
sequence $(G_d)_{d \ge d_0}$ of subgroups of $\Delta S_d$ is \emph{natural} 
if for some positive integer $p < d_0$,
(a) $i_{d,p}(G_d) \subset  G_{d+1}$, and (b) $\text{dim}(M(k,d)^{G_d})$ is 
independent of $d \ge d_0$ (assume $k \ge d$).
\begin{example}\label{ex: fam}
Set $G_d = \Delta (S_{d-p} \times  S_p)$ and $m$ be a positive 
integer.  If $p=0$, then $G_d = \Delta S_d$ and  
$\text{dim}(M(d+m,d)^{G_d})= 2+m$, $d \ge d_0=2$; if $p=1$, then  
$\text{dim}(M(d+m,d)^{\Delta (S_{d-1} \times S_1)}) = 5+2m$,  $d \ge 3$;
if $p\ge 2$, then $\text{dim}(M(d+m,d)^{\Delta (S_{d-p} \times S_p)}) = 
6+2m$, $d \ge 4$. 
\end{example}

If $(G_d)_{d \ge d_0}$  is natural, we often identify $M(k,d)^{G_d}$ with 
$\real^N$, $d \ge d_0$, where $\text{dim}(M(k,d)^{G_d})=N$. We define 
linear isomorphisms $\tXi:\real^N \arr M(k,d)^{G_d}$ for the families of 
\pref{ex: fam}. Let $I^\star_i = \mathcal{I}_{i,i} - 
I_i,~i\in\pint$. The matrix $\tXi(\tbxi)$, $\tbxi\in\real^N$, is expressed 
as block 
diagonal matrix $[B_{ij}]$, where each $B_{i,j}$ is a linear combination in 
the 
coordinates of $\tbxi$ of the matrices 
$\mathcal{I}_{i,j}$, $I_i$ and $I^\star_i $. 
For example, if $p = 1$, $ m \ge 1$, then $N = 5+2m$ and 
$\tXi(\txi_1,\cdots,\txi_N)$ 
is the $(2+m) \times 2$-block matrix $[B_{ij}] \in M(d+m,d)^{\Delta 
(S_{d-1} \times S_1)}$ defined by
\begin{align*}
B_{11}&= \txi_1 I_{d-1} + \txi_2 I^\star_{d-1},\; B_{12} = \txi_3 
\mathcal{I}_{d-1,1}, \\
B_{i1}&= \txi_{2i} \mathcal{I}_{1,d-1}, \;B_{i2}=  \txi_{2i+1} 
\mathcal{I}_{1,1}, \; 2 \le i \le 2+m.
\end{align*}
Similar expressions hold for the other families.  In practice, we restrict 
the vector field $\nabla \ploss$ to $M(k,d)^{G_d}$ and then pull back this 
vector field using $\Xi$ to a vector field $\vectfieldidx{d}$ on $\real^N$. 
The 
Jacobian of $\vectfieldidx{d}$ is then equal to the Hessian of $\nabla 
(\ploss 
| M(k,d)^{G_d})$. Observe that $\vectfieldidx{d}$ does not depend on a 
choice of inner product on $\real^N$; indeed, we take the standard 
Euclidean inner product on $\real^N$ ($\Xi: \real^N \arr M(k,d)^G$ is 
\emph{not} an isometry). Since $\vectfieldidx{d}(\tbxi) = \tXi^{-1} \nabla 
\ploss(\tXi(\tbxi))$, we may read off the components of 
$\vectfieldidx{d}(\tbxi)$ directly from the corresponding matrix entries 
of $\nabla \ploss$. We find that $\vectfieldidx{d}$ is a continuous 
family of vector fields on 	$\real^N$ in the \emph{real} parameter $d$. 
Obviously, no such statement can hold on $M(k,d)$ as the dimension of 
$M(k,d)$ depends on $d$.
Moreover, the vector fields $\vectfieldidx{d}$ will be real analytic 
outside of 
a thin 	(semianalytic) subset of $\real^N$. Indeed, $\vectfieldidx{d}$ is 
subanalytic~\cite{bierstone1998subanalytic} 
but the real analyticity statement is easily proved 
directly and suffices for our applications.  All of this allows us to 
``connect'' the critical 
points $\mathfrak{c}(d)\in M(k,d)^{G_d}$, $d \ge d_0$, by curves in 
$\real^N$ and develop a path-based approach to our problem.  
\begin{definition}\label{def: Efamily}
(Notation \& Assumptions as above.) A family 
$\mathfrak{C} = \{\mathfrak{c}(d)\dd d \ge d_0\}$ of critical points of 
$\cL$ with isotropy $G_d \subset \Delta S_d$ is \emph{weakly regular} 
if $(G_d)_{d \ge d_0}$ is natural and for $d \ge d_0$ 
\begin{itemize}
	\item[(a)] There is a continuous curve $\gamma_d:[0,1]\arr \real^N$ 
	of critical points of $\vectfieldidx{d}$ joining $\mathfrak{c}(d)$ 
	to 
	$\mathfrak{c}(d+1)$.
	\item[(b)] $\ploss$ is real analytic at $\mathfrak{c}(d)\in 
	M(k,d)$.  
	\item[(c)] $\vectfieldidx{d}$ is real analytic on a neighbourhood 
	of 
	$\gamma_d([0,1]) \subset \real^N$. 
	\item[(d)] The Jacobian of $\vectfieldidx{d}$ along $\gamma_d$ is 
	non-singular. 
\end{itemize}
If $\,\lim_{d \arr \infty} \mathfrak{c}(d) \defoo 
\mathfrak{c}_\infty\in\real^N$ exists and is bounded (Euclidean norm on 
$\real^N$), the family is \emph{regular}. 
\end{definition}
Using (c,d), and the real analytic implicit function theorem, $\gamma_d$ is 
real analytic. 
The family defined by $\mathfrak{c}(d) = V$ is not weakly regular as (b) 
fails even if $k = d$.

It may be shown, using results on subanalytic sets and the Curve 
Selection Lemma \cite{john1968singular}, that every regular family 
has a fractional power series (FPS) representation. That is, with 
the notation and assumptions of \pref{def: Efamily}, there exist 
$d_1 \ge d_0$ and a minimal $\base 
\in \pint$ such that each component $\mathfrak{c}_i(d)$ of 
$\mathfrak{c}(d) \in \real^N$ is given by the 
convergent power series

\begin{align} \label{eqn:fps_rep}
\mathfrak{c}(d)_{i} = \sum_{j=0}^\infty \fpscoeff{i}{j}  
d^{-\frac{j}{\base}},\; i
\in [N].
\end{align}
Under the assumption of weak regularity, there may be Puiseux series
representations~\cite[Exam.~4.13]{ArjevaniField2020} and 
$\mathfrak{c}_\infty$ lies in the one point compactification of $\real^N$. 
In practice, rather than use the general result, we prove directly that a 
family has an FPS representation. Verifying regularity for sufficiently 
large $d_0$, is usually straightforward or trivial.  We refer to 
\pref{sec:family_pf} for examples of construction of FPS 
representations when~$k>d$.

The FPS representation for families of critical points is important 
both theoretically, and computationally and yields Puiseux series 
representations of the objective value and Hessian spectrum. It was shown 
in \cite[Section 8]{ArjevaniField2020} that for $k = d$ 
several (regular) families of
critical points with isotropy $G_d = \Delta(S_{d-p} \times S_p)$, $p 
\in \{0,1\}$, had FPS representations in $d^{-\frac{1}{2}}$ (so 
$\base = 2$). 
Each coordinate of $\mathfrak{c}_\infty = \lim_{d \arr \infty} 
\mathfrak{c}(d)\in \real^N$ was either $\pm 1$ 
or zero. For examples of FPS representations with $k = d$ and $G_d = 
\Delta 
(S_{k-p} \times S_p)$, $p \in \{2,3\}$, see 
\cite{arjevanifield2021analytic}. 
Explicit construction of the coefficients in these FPS examples is 
relatively straightforward and algebraic formulae can be 
given for low order terms. When $k > d$, analysis is harder. It is 
not always possible to give low order coefficients in a simple 
algebraic form.  Moreover, there may be multiple regular families 
with the same limiting value $\mathfrak{c}_\infty \in \real^N$.

\begin{definition}\label{def: types}
Let $p \ge 0$ and take $G_d = \Delta(S_{d-p} \times S_p)$ as in 
\pref{ex: fam}. A regular family of critical points with 
isotropy $(G_d)_{d \ge d_0}$ is of type I (resp.~type II)
if as $d \arr \infty$, the diagonal elements of the $(d-p)\times 
(d-p)$-block corresponding to the action of $\Delta 
S_{d-p}$ converge to $-1$ (resp.~$+1$).\\
\end{definition}
In terms of FPS, a family is of type I (resp.~type II) if 
$\fpscoeff{1}{0} = -1$ (resp.~$+1$).
\begin{rem}\label{rem:halfway}
When $k = d$, we suspect that the initial coefficients 
$\fpscoeff{i}{0}$ of the FPS of a regular family with isotropy $\Delta 
(S_{d-p} \times S_p)$, 
$p\ge 0$, 
satisfy $\fpscoeff{i}{0} \in \{\pm 1, 0\}$. However, this is false 
if $k > d$---we give an example later. Moreover if $k=d$, and with a 
slight extension of the notion of regular family, there is a regular 
family with $G_{2d} = \Delta (S_d \times S_d)$, $d \in \pint$, with 
$\fpscoeff{1}{0} = +1$, $\fpscoeff{5}{0} = -1$ (coefficients 
corresponding to the diagonal entries of the principal blocks 
associated to $S_d \times {I_d}$ and $I_d \times S_d$).
\end{rem}

Henceforth, we emphasize $k > d$, and families with isotropy $\Delta 
(S_{d-p} \times S_p)$, $p\in\{0,1\}$.  However, the methods are 
quite general. 
\begin{theorem} \label{thm:only_two}
Suppose that $\mathfrak{C}$ is a regular family of critical points. Assume  
that initial terms of the asociated FPS do not all vanish and 
$\base\in\{2,4\}$. If the isotropy $G_d = \Delta S_d$, $k \in \{d,d+1\}$, then 
$\mathfrak{C}$ is of type~I; if the isotropy $G_d = \Delta (S_{d-1} \times 
S_1)$, $k \in \{d,d+1,d+2\}$, then $\mathfrak{C}$ is either of type~I or 
type~II and there exists at least one family of each type. If $k = d$ there is 
precisely one type I family and, if $p \ne 0$, one type II family.
\end{theorem}


\begin{rem}\label{rem:Xavier}
For $k = d$, both type I and type II families are spurious 
minima~\cite{arjevanifield2020hessian,arjevanifield2021analytic}. 
However, empirically, type I minima are not detected by SGD when Xavier 
initialization is used. Since the loss at type II 
minima 
decays as
$\Theta(1/d)$ and the loss at
type I is $\Theta(1)$ (independently of the isotropy), it may be
tempting to argue that the expected initial loss under Xavier 
initialization	is smaller than the loss at type I minima. However, 
this turns out to be false: Assume $k = d$. Under Xavier 
initialization,	$\prn*{1-\frac{2}{\pi}}d\le E_{W}[\cL(W)] \le 
\prn*{1-\frac{1}{\pi}}d$ (see \pref{sec:expiv}). 
\end{rem}

As we increase $k-d$, the original type I and II critical points of 
spurious minima persist as degenerate critical sets and new regular 
families of critical points of the same type are generated. Thus, if $k - d 
= 1$, and $G_d \subsetneq \Delta S_d$, 
two regular families of type I points are generated which are swapped by 
the permutation of rows $d$ and $d+1$.  
Similarly for families of type II. When $k-d=2$,  $3! = |S_3|$ new regular 
families of type I critical points appear; similarly for type II. 
Additional families of critical points, which do not originate from the 
original families and are not spurious minima, may appear. 
See \pref{sec:fossil} for degenerate critical point sets 
occurring on account of over-parameterization.

Our focus will be on the families of type I and II critical points that 
arise through the above mechanism.
For a given isotropy $G_d = \Delta (S_{d-p}\times S_p)$, $p \in \{0,1\}$, a 
type $X$ and $k \ge d$, with $m = k-d$ fixed, let $\bigfamily{X}{p}{m}$
denote a choice of regular family of critical points 
$\{\smallfamily{X}{p}{m}(d)\in M(k,d)^{G_p}\}$ that 
originates from the unique regular family of critical points of type X that 
exists when $k = d$.  It is 
enough to analyze just one of the type X families when $k > d$ as, 
by equivariance, the choices lie on the same $S_d \times S_k$-orbit 
and so have similar Hessians. Once the existence of the FPS 
representation for the families 
$\bigfamily{X}{p}{m}$ has been proved, the next step is to estimate the Hessian 
spectrum, the topic of the next~section.

We conclude with examples illustrating the quantitative power of our 
approach. 

\begin{example}\label{ex: fpsapp}
(1) We investigate how the loss $\ploss(\mathfrak{c}^{II}_{1,m})$ 
depends on $k - d$ for the type II families $\bigfamily{II}{1}{m}$, $m 
\in \{0,1,2\}$.  For $m > 0$, the initial coefficients of the FPS are 
found using Newton-Raphson method applied either directly for a system of 
equations corresponding to different orders of the FPS coefficients 
\pref{sec:typeIorII}, or for a reduced system of equations obtained through an 
explicit use of the geometry of the problem \pref{sec:fps_der}.
We give the asymptotics modulo 
$O(d^{-\frac{3}{2}})$ and find that if
$\ploss(\mathfrak{c}^{II}_{1,m}) = \alpha_m d^{-1} + 
O(d^{-\frac{3}{2}})$, then
\[
\alpha_0 = 2.97357632715\ldots (=\frac{1}{2} - \frac{2}{\pi^2}), \; 
\alpha_1 = 2.67254813889\ldots, \; \alpha_2 = 2.67193392202\ldots
\]
(2) Consider $\ploss(\mathfrak{c}^{I}_{1,m})$ for type I 
families. We 
find that $\ploss(\mathfrak{c}^{I}_{1,m}) = \frac{1}{2} - \frac{1}{\pi} 
+ O(d^{-\frac{1}{2}})$, for $m\in  \{0,1,2\}$.
Higher order terms are $m$-dependent but can be computed, as they can 
if $p = 0$. For example, if $p = 0$, $m = 1$ then 
$\ploss(\mathfrak{c}^{I}_{0,1})=\frac{1}{2} - \frac{1}{\pi} - 
\frac{4}{3 \pi} d^{-\frac{1}{2}} + \prn*{-1 - \frac{2}{\pi^{2}} + 
	\frac{4}{\pi}}d^{-1} + O(d^{-\frac{3}{2}})$.\\
\noindent(3) We conclude with an example where the FPS is in powers of 
$d^{-\frac{1}{4}}$ and two components of 
$\mathfrak{c}_\infty\in\real^7$ do not lie in $\{\pm 1,0\}$. If $k = 
d+1$, the initial terms of the FPS for a type I family 
$\mathfrak{c}(d)$ of 
critical points of spurious minima are given by 
\begin{align*}
	\mathfrak{c}(d)_1& = -1 + 2d^{-1} + \frac{\pi}{2} d^{-\frac{3}{2}} 
	+ O(d^{-\frac{7}{4}}),\hspace*{0.15in}\mathfrak{c}(d)_2= 2d^{-1} - 
	\sqrt{\pi-2}d^{-\frac{7}{4}}+ O(d^{-2}),  \\
	\mathfrak{c}(d)_3 & = d^{-1} - 
	\frac{6+3\pi}{4\pi\sqrt{\pi-2}}d^{-\frac{5}{4}}+ 
	O(d^{-\frac{3}{2}}), \;\mathfrak{c}(d)_4= \frac{\sqrt{\pi-2}}{2} 
	d^{-\frac{3}{4}} + O(d^{-1}) ,\\
	\mathfrak{c}(d)_5 & = \frac{1}{2} + 
	\frac{6+3\pi}{8\pi\sqrt{\pi-2}}d^{-\frac{1}{4}} + 
	O(d^{-\frac{1}{2}}),\hspace*{0.18in} \mathfrak{c}(d)_6= 
	\frac{\sqrt{\pi-2}}{2}d^{-\frac{3}{4}} + O(d^{-1}),\\
	\mathfrak{c}(d)_7 & = -\frac{1}{2} + 
	\frac{6+3\pi}{8\pi\sqrt{\pi-2}}d^{-\frac{1}{4}} + 
	O(d^{-\frac{1}{2}})
\end{align*}
(see \pref{sec:typeIorII} for details). The loss 
$\ploss(\mathfrak{c}^{I}_{1,1}) 
= \frac{1}{2} - \frac{1}{	\pi} - \frac{4}{3\pi 
}d^{-\frac{1}{2}} - \frac{(\pi -2)^{\frac{3}{2}}}{3\pi} 
d^{-\frac{3}{4}} 
+ O(d^{-1})$.
\end{example}

\section{Hessian spectrum} \label{sec:hes_spec0}
The FPS representation makes possible an analytic characterization of the 
Hessian spectrum 
using tools from the representation theory of groups (see
\pref{sec:hes_spec_pf} for a brief review). The main tool used 
is the	\emph{isotypic decomposition} relating the isotropy of a given 
point to minimal invariant subspaces of the Hessian. We begin by 
presenting the isotypic decomposition needed for the 
over-parameterized~case.

Let $k \ge d$. Regard $M(k,d)$ as an $S_d$-representation (diagonal action 
on $M(d,d) \subset M(k,d)$). 
By restriction, $M(k,d)$ is a $S_q \times S_p$-representation, where $q = 
d-p$, $p < q$, and $S_{q} \times S_p \subset S_d$. If 
$p \in \{0,1\}$ (the case of interest here)
the isotypic decomposition uses 4 irreducible representations of $S_{q}$, 
when $d \ge 4$: the trivial representation
$\mathfrak{t}$ of degree 1, the standard representation $\mathfrak{s}_{q}$ 
of
$S_{q}$ of degree $q-1$, the exterior square representation
$\mathfrak{x}_{q}=\wedge^2 \mathfrak{s}_{q}$ of degree 
$\frac{(q-1)(q-2)}{2}$ and a
representation $\mathfrak{y}_{q}$ of degree $\frac{q(q-3)}{2}$
(associated to the partition 
$(q-2,2)$~\cite{James1978,fulton1991representation}).
For $p  \in \{0,1\}$,  the isotypic decomposition is
\begin{align}
M(k,d) = (m + 3p + 2)\mathfrak{t} + (pm + 2p + 
3)\mathfrak{s}_{q} 
+               \mathfrak{x}_{q} + \mathfrak{y}_{q}.
\end{align}

Since the representations $\mathfrak{x}_{q}, \mathfrak{x}_{q}$ contribute 
$2$ eigenvalues, of total multiplicity $q^2-2q+1$, we have
\begin{lemma} \label{lem:dis_evs} If $k-d = m\ge 0$ and $p \in \{0,1\}$, 
then of the $kd$ eigenvalues of the Hessian at a point of isotropy $\Delta 
(S_{d-p}\times S_p)$:
\begin{enumerate}
	\item $\Theta(d^2)$ are populated by two eigenvalues: the $\fx$-
	and $\fy$-representation eigenvalues.
	\item At most $O(k-d)$ eigenvalues are distinct.
\end{enumerate}
\end{lemma}
\pref{lem:dis_evs} implies that the isotropy type of a 
point strictly restricts the number of distinct eigenvalues of the Hessian 
spectrum. For fixed $k-d$, the $\fx$-	and the $\fy$-representation eigenvalues
account for $kd - \Theta(d)$ of the eigenvalues. We show that for
all families of critical points considered here, the $\fx$-
and the $\fy$-representation eigenvalues are identical to order 
$O(d^{-\frac{1}{4}})$.

\begin{theorem} \label{thm:xy_evs}
For a family of critical points of isotropy $\Delta(S_{d-p}\times 
S_p),~p\in\{0,1\}$ and $k$ as in \pref{thm:only_two}, 
$kd - \Theta(d)$ of the Hessian eigenvalues are
populated by the two eigenvalues: $$\frac{1}{4} - \frac{1}{2\pi} + 
O(d^{-\frac{1}{4}})\quad 
\text{and }\quad \frac{1}{4} + \frac{1}{2\pi} + O(d^{-\frac{1}{4}})$$
associated to the $\fx$- and the $\fy$-representation, respectively.
\end{theorem}

The derivation of \pref{thm:xy_evs} builds on a technique used
in \cite{arjevanifield2020hessian,arjevanifield2021analytic} and is directed 
towards the case where $k > d$ and the coefficients of FPS may not be given in 
a simple algebraic form. Specifically, we rewrite the expression for the 
Hessian eigenvalues in terms of the gradient entries. Since gradient entries 
vanish at critical points, this allow us to evaluate the eigenvalue 
expressions. For example, 
the Puiseux series of the $\fx$-eigenvalue of Type I 
$\Delta (S_{d-1}\times S_1)$-critical points is
\begin{align}\label{eqn: x_relation}
	\lambda_\fx^d &= 
	\frac{1}{4} - \frac{1}{2 \pi} + [d^0]\partder{1} - 
	c_{1,2}[d^{\frac{1}{2}}]\partder{1} - [d^0]\partder{2}+
	[d^{\frac{1}{4}}]\partder{1} d^{\frac{1}{4}}+
	[d^{\frac{1}{2}}]\partder{1} d^{\frac{1}{2}} + O(d^{-\frac{1}{4}}), 
\end{align}
with $[d^\alpha] \partder{i}$ indicating the coefficient of $d^\alpha$ in 
$\partder{i}$. Since $\partder{i}$ vanish at critical points, 
$\lambda_\fx^d = \frac{1}{4} - \frac{1}{2 \pi} + O(d^{-\frac{1}{4}})$. 
\pref{eqn: x_relation} 
further demonstrates the sensitivity of the $\fx$-eigenvalue to 
variations in the FPS coefficients and in different orders of 
the gradient terms, see \pref{sec:xy_evs_pf}. Algebraic relations between 
criticality and curvature indicate therefore a certain rigidity of the loss 
landscape. Relations of similar nature exist between 
\emph{criticality} and the \emph{loss} at a point. 

It follows from \pref{thm:xy_evs} that all Hessian eigenvalues not 
associated to the trivial or standard representations are strictly positive 
for sufficiently large $d$. Consequently, annihilation of spurious minima in a 
family must be tangent to an invariant subspace of the sum of the isotypic 
components for the trivial and standard representations. Generically, it is 
to be expected that the subspace will be isomorphic to either the trivial 
representation or the standard representation.

\section{Over-parameterization} \label{sec:over}
Having computed the $\mathfrak{x}$- and the $\mathfrak{y}$-eigenvalues, we now 
turn to describe how the eigenvalues associated to the trivial and the standard 
representations vary when the number of neurons is increased. We find that 
while eigenvalues associated to the trivial representation remain strictly 
positive for all sufficiently large~$d$---some eigenvalues, associated 
to the standard representation, become negative, indicating a transition from 
minima to saddles along the isotypic component of the standard representation. 
We start with points of isotropy~$\Delta S_d$.

\subsection{Critical points of isotropy $\Delta S_d$}
By \pref{thm:only_two}, if $k \in \{d,d+1\}$, there is one regular 
family of critical points with isotropy 
$\Delta S_d$: the type I family $\bigfamily{I}{0}{i}$, $i =k-d$. 
The representation-theoretic tools used in \pref{sec:hes_spec0}, 
yield a complete characterization of the Hessian spectrum of both families 
of critical points (see the discussion following the statement of 
\pref{thm:main0} in the introduction for more details).

The spectral analysis of the Hessian reveals that: 
\begin{enumerate}[leftmargin=*,label=\Alph*.]
\item $\bigfamily{I}{0}{0}$ is a family of
minima.	
\item Adding one neuron turns it into the family
$\bigfamily{I}{0}{1}$
of non-degenerate saddles where the negative eigenvalue of the Hessian at 
$\bigfamily{I}{0}{1}$ is
associated to the standard representation $\mathfrak{s}_d$. 
\end{enumerate}
Since the negative eigenvalue of the Hessian at $\bigfamily{I}{0}{1}$ is 
associated to $\mathfrak{s}_d$, there are exactly $d-1$ descent 
directions, out of $d(d+1)$ possible directions in $M(d+1,d)$, lying in the 
$4d-4$-dimensional isotypic component $4\mathfrak{s}_{d}$ spanned by (only 
nonzero elements are described):
\begin{enumerate}[leftmargin=*]
	\item The $(d-1)$-dimensional space of $(d+1)\times d$-matrices 
	$[y_{ij}]$ where for $i,j \in \is{d}$, $y_{ij} = z_i - z_j$,
	for some $(z_1,\cdots,z_d) \in \real^d$ with $\sum_{i \in \is{d}} z_i = 
	0$. 
	
	\item The $(d-1)$-dimensional space of $(d+1)\times 
	d$-matrices $[y_{ij}]$ where for $i,j \in 
	\is{d}$, $i \ne j$, $y_{ij} = z_i + z_j$, where $(z_1,\cdots,z_d) \in 
	\real^d$ with $\sum_{i \in \is{d}} z_i = 0$. 
	
	\item The $(d-1)$-dimensional space of $(d+1)\times 
	d$-matrices whose diagonal elements sum up to zero.
	
	\item The $(d-1)$-dimensional space of $(d+1)\times 
	d$-matrices whose $(d+1)$'th row elements sum up to zero.
\end{enumerate}
No regular families with isotropy $\Delta S_d$ 
exists if two neurons or more are added, i.e., $k - d \ge 2$ (see 
\pref{sec:fossil}).

\subsection{Critical points of isotropy $\Delta (S_{d-1}\times S_1)$}
Consider the type II regular families
$\bigfamily{II}{1}{0}$,
$\bigfamily{II}{1}{1}$ and $\bigfamily{II}{1}{2}$, corresponding to $k=d, 
d+1$ 
and $d+2$ respectively.
We show that negative eigenvalues of the Hessian appear when $k-d = 2$ but 
not $k - d = 1$. The same result (and proof) hold for the type I family.

For $k=d, k=d+1$, the eigenvalues associated to the trivial and 
standard representation $\mathfrak{s}_{d-1}$ are strictly positive---see 
\pref{table:typeIIevs}.
By \pref{thm:xy_evs}, the $\fx$- and the
$\fy$-representation eigenvalues are also strictly positive. Therefore,
we have families of spurious minima for $k = d,d+1$. 
\begin{table}[h]\label{table:typeIIevs}
\begin{center}
	\begin{tabular}{ |c||c|l|l|c } \hline
		Isotypic component   &Degree & 
		\hspace*{0.3in}$k=d$&\hspace*{0.2in}$k =d+1 $\\\hline
		&&Symmetry $\Delta S_{d-1}$&Symmetry $\Delta 
		S_{d-1}$\\\hline\hline
		Trivial representation & 1 & 0.0908 & 0.0044,  0.0843\\
		&&0.25& 0.2632, 0.3121 \\
		mult.~$5$, if $k=d$&&0.1591$d$ - 0.3471 & 0.1591$d$ + 0.7546\\
		mult.~$7$, if $k=d+1$          &&0.25$d$ + 0.25& 0.25$d$ + 0.5\\
		&&0.25$d$ + 0.8471 & 0.25$d$ + 1.0979\\
		
		\hline
		Standard representation $\mathfrak{s}_{d-1}$ & $d-2$ & 0.0908 & 
		0.0230, 0.0908 \\
		&& 0.0908  & 0.0936\\
		mult.~$5$, if $k=d$         && 0.25 &0.2693\\
		mult.~$6$, if $k=d+1$ && 0.4091 & 0.5340\\
		&& 0.25$d$ + 0.25& 0.25$d$ + 0.5\\
		\hline
		Loss &&         $\Theta(1/d)$ &$\Theta(1/d)$
		\\
		\hline
	\end{tabular}
\end{center}
\caption{Type II critical points with symmetry $\Delta S_{d-1}$. The 
	Hessian eigenvalues  associated to the trivial and the
	standard representation are given to four decimal places, 
	modulo
	$O(d^{-\frac{1}{2}})$-terms.
}
\end{table}
For $k=d+2$, critical points in the family $\bigfamily{II}{1}{2}$ 
are saddles, with strictly negative	eigenvalues associated to 
$\mathfrak{s}_{d-1}$. To show this, we take a different route so as to reduce 
the 
complexity of the computation. Consider the $2d\times 2d$-submatrix 
$\widehat{H}$ of the Hessian  corresponding to the last two rows of the 
weight matrix. By Cauchy's interlacing theorem \cite{johnson1985matrix}, 
the smallest eigenvalue of the Hessian is bounded from above by the 
smallest eigenvalue of $\widehat{H}$. Therefore, it suffices to prove that 
$\widehat{H}$ has a negative eigenvalue. Since the isotypic decomposition 
corresponding to $\widehat{H}$ consists of exactly two of the subspaces 
associated to $\mathfrak{s}_{d-1}$, 
the problem is reduced to computing the spectrum of a $2\times 2$
matrix. Using Puiseux series representation, we show that modulo
$O(d^{-\frac{1}{2}})$-terms the two eigenvalues of $\hat{H}$ are 
$\lambda_1 = 
0.8060\ldots$ and $\lambda_2=-0.1198\ldots$. Hence 
there exists a $(d-2)$-dimensional eigenspace of descent
directions, projecting onto the associated eigenspace for $\widehat{H}$.
Applying Cauchy's interlacing theorem again,  there must also exist
positive eigenvalues associated to $\mathfrak{s}_{d-1}$. 
\\
\begin{theorem}\label{thm:final} (Notation \& Assumptions as above.) 
\begin{itemize}
	\item $\Delta S_{d}$-symmetric critical points of type I are minima 
	for $k=d$ and non-degenerate saddles for $k=d+1$ with a 
	$(d-1)$-dimensional eigenspace of descent directions.
	\item 
	Critical points of isotropy $\Delta (S_{d-1} \times 			
	S_1)$, type I or II, define regular families of	spurious minima for 
	$k=d,d+1$, and non-degenerate saddles for $k=d+2$, with at least a 
	$(d-2)$-dimensional space of descent directions.
\end{itemize}
\end{theorem}

Empirically, when $k=d+1$, minima of isotropy $\Delta (S_{d-1} \times S_1)$ 
are not seen for $d < 8$ (they are if $k = d$ and $d \ge 6$) and the 
probability to detect them using gradient descent is 
much lower for small values of $d\ge 8$~\cite{safran2017spurious} (type I 
minima are not detected under Xavier 
initialization). \pref{thm:final} implies that when $k = d+2$ there 
are no spurious minima of type II of symmetry $\Delta( 
S_{d-1} \times S_1)$ and the descent directions are tangent to a copy of 
$\mathfrak{s}_{d-1}$. 
The last point is crucial for understanding the empirical results. The 
small 
eigenvalue $0.0230$ associated to $\mathfrak{s}_{d-1}$ when 
$k = d+1$
indicates that we are close to a change of stability (bifurcation) of the 
critical point for gradient descent. Bifurcation of the trivial solution on 
$\mathfrak{s}_{d-1}$
is special and quite exceptional. 
In our case, the trivial solution will be a sink for 
gradient descent (i.e., a strict local minimum of the loss function), when 
$k = d+1$, and a source (i.e., a strict local maximum of the loss function) 
when $k=d + 2$. The change of stability results from the collision of a 
large number of saddles of \emph{high index} with the sink, followed by the 
emergence of a source and a large number of saddles with low index.	The 
high index of the saddles converging to the sink, implies that the basin of 
attraction for the sink shrinks rapidly as the saddles approach the sink. 
We refer to	\cite[Sections 1.1, 4]{arjevanifield2022equivariant} for more 
on this phenomenon. As we increase $d$, families $\bigfamily{II}{p}{ }$, $ p > 
1$, of spurious minima appear which may not be annihilated by adding two 
neurons.  However, no such minima were found in \cite{safran2017spurious} 
when $k = d+2, d \le 20$ (they were for $k = d+1$).	The empirical results 
provide strong support for a change of stability associated to 
$\mathfrak{s}_{d-1}$ and suggest the unique role this representation may 
play in understanding over-parametrization.

\section*{Concluding Comments}
The rich symmetry structure exhibited by the loss function 	
(\ref{opt:problem}) makes possible an analytic study of the associated 	
nonconvex loss landscape. The approach is twofold. First, the presence 	
of symmetry breaking allows an efficient organization of an	otherwise 
highly complex set of critical points, offering new ways of recognizing, 
differentiating and understanding local minima (\pref{sec:sym_break} and 
\pref{sec:family}). Second, for a given family of critical points, symmetry 
grants a parameterization of fixed dimensionality, independent of the ambient 
space, which permits a new array of analytic and algebraic tools 
(\pref{sec:hes_spec0} and \pref{sec:over}).

In this work, the symmetry breaking framework is used for investigating the 
nature by which overparameterization contributes to making the 
loss landscape of (\ref{opt:problem}) accessible for gradient-based 
methods. We find that increasing the number of neurons transforms spurious 
minima into saddles: decent directions are formed along linear subspaces 
corresponding the standard representation $\mathfrak{s}_d$ of $S_d$, ascent 
directions along other representations of $S_d$ persist, and the loss remains 
essentially the same. The process by which spurious minima turn into saddles 
suggests a powerful mechanism enabling minimization of the~\emph{loss} (rather 
than the gradient norm 	
\cite{ghadimi2013stochastic,nesterov2006cubic,arjevani2022lower}) 
via computationally efficient local search methods, and further highlights 
the importance of the intricate interplay between symmetries inherent to 
data distributions and those displayed by neural network models.

Our spectral results assume the target $\VV$ has high symmetry but apply also 
to asymmetric problems. In the cases we discuss, the transition from saddle 
to minimum, or minimum to saddle, occurs at a non-integer value of $d$. 
Hence, at integer values of $d$, critical points are non-degenerate.
It follows that for the families $\bigfamily{X}{p}{m}$ we consider, there is an 
open neighborhood $\mathcal{U}$  of $V = I_d\in M(d,d)$, such	that for all 
$V' \in \mathcal{U}$, the loss function $\ploss'$ for $V'$ has a non-degenerate
critical point close to each point of the $\Delta (S_{d-p}\times 
S_p)$-orbit of $\bigfamily{X}{p}{m}$ with the same number of negative 
eigenvalues (counting multiplicities). Critical points and eigenvalues 
depend continuously on $V'\in\mathcal{U}$ and  the Hessian spectrum remains 
extremely skewed. 	

There is the problem of understanding the geometric mechanisms underlying 
the transition from minimum to saddle. As already indicated, this is 
closely related to the geometry of the standard representation 
$\mathfrak{s}_d$ of $S_d$. For simplicity, assume $d=2\ell+1$ is odd 
(similar results hold if $d$ is even~\cite{arjevanifield2022equivariant}).
Gradient vector fields on $\mathfrak{s}_d$ always have a critical point at 
the origin which is never a non-degenerate saddle but is (generically) 
either a non-degenerate minimum or maximum. The transition between minima 
(sink for the gradient descent, index $2\ell$) and maxima (source, index 
$0$) can be achieved \emph{locally} (that is, as a local deformation of the 
landscape geometry) by $2^{d-1}-1$ non-degenerate saddles of index $\ge \ell$
passing simultaneously through the origin and emerging as $2^{d-1}-1$ 
non-degenerate saddles of index $\le \ell$. \emph{No new minima or maxima 
are created}. Forced symmetry breaking leads to great complexity near the 
transition but minimal models of complexity can be given (\emph{op.~cit.}, 
Section 4).

Rather than striving for generalization, our approach in this work has been 
to focus on an analytically tractable case, one already acknowledged as 
difficult 
\cite{chizat2018global,mei2018mean,goldt2019generalisation,
brutzkus2017sgd,safran2017spurious,jacot2018neural}, that helps 
elucidate some of the key foundational questions. The phenomena described 
are robust and so already have the power to disprove or support general 
conjectures in DL 	
\cite{ArjevaniField2020,arjevanifield2020hessian,arjevanifield2021analytic}.
The symmetry breaking framework used to study these phenomena 
generalizes beyond the families of minima considered in the present work 
\cite{arjevanifield2022bifurcation}, and applies to other choices of activation 
functions and distributions \cite{arjevanifield2019spurious,arjevanifield2021tensor}.
In addition, numerical work indicates that minima of the
empirical loss are also symmetry breaking, and so allow theoretical 
investigations of the empirical loss landscape as well as algorithmic biases 
(see, e.g., \cite{gunasekar2018characterizing}) within the new analytic 
framework. The full scope and power of symmetry breaking in DL, and more 
generally stochastic nonconvex optimization, remain to be~discovered.

\subsection*{Acknowledgments}
We thank Noa Aharon, Daniel Soudry and Avi Wigderson for valuable 
discussions and constructive suggestions. The research was supported 
by the Israel Science Foundation (grant No. 724/22).


\bibliographystyle{unsrt}
\bibliography{bib}

\newpage
\appendix


\section{Supplementary material for 
\pref{sec:family}}\label{sec:family_pf}
\newcommand{\defox}{\stackrel{\mathrm{def}}{=}}
We begin by outlining the proof of existence for type II families with $k = 
d+2$ and isotropy $\Delta (S_{d-1}\times S_1)$ following the strategy in 
\cite[Section 8]{ArjevaniField2020}, and briefly discuss other cases given in 
\pref{thm:only_two}. Estimates required 
for a valid use of the real analytic 
version of the implicit function theorem are obtained using two different 
methods presented in \pref{sec:fps_der} and in \pref{sec:typeIorII}.

\subsection{Fractional power series representation for type II minima: $k = 
d+2$} \label{sec:fps_der}
Following~\pref{sec:family}, we restrict the loss $\cL$ to the fixed point space $M(d+2,d)^{\Delta S_{d-1}}$, which is of dimension $9$ for $d \ge 3$. We recall the linear isomorphism
$\tXi: \real^9 \arr M(d+2,d)^{\Delta S_{d-1}}$ defined by
\[
\tXi(\tbxi) =
\left[\begin{matrix}
                        \txi_1\cI_{d-1,d-1} + \txi_2I_{d-1}^\star & \txi_3\cI_{d-1,1}\\
                        \txi_4\cI_{1,d-1} & \txi_5\cI_{1,1}\\
                        \txi_6\cI_{1,d-1} & \txi_7\cI_{1,1} \\
                        \txi_8\cI_{1,d-1} & \txi_9\cI_{1,1}
\end{matrix} \right], \;\tbxi = (\txi_1,\cdots,\txi_9)\in \real^9
\]
Regard $\tXi$ as an identification of $M(d+2,d)^{\Delta S_{d-1}}$ with $\real^9$ and recall from \pref{sec:family} 
that $\vectfieldidx{d}$ is the vector field on $\real^9$ defined as 
the 
pullback of $\nabla \ploss| M(k,d)^{\Delta S_{d-1}}$ by
$\Xi$.

The map $\tXi$ naturally determines a $4 \times 2$-block 
decomposition of matrices in $M(d+2,d)^{\Delta S_{d-1}}$.  
If a row $\w$ of $W\in M(d+2,d)^{\Delta S_{d-1}}$ lies in 
row $\ell$ 
of the block decomposition, the row is said to be of \emph{row-type} $\ell$. Necessarily $\ell\in [4]$. Clearly there are $d-1$-rows of row-type 1 and exactly one row of row-type $\ell$, $2 \le \ell \le 4$.

\begin{rem}
	The vector fields $(\grad{\ploss})| M(k,d)^G$, $\nabla (\ploss|M(k,d)^G)$
	on $M(k,d)^G$ are equal, \emph{provided that we use the inner product on
		$M(k,d)^G$ induced from $M(k,d)$ to define the gradient of $\ploss |
		M(k,d)^G$}, and so the eigenvalues of the Hessian of $\ploss$ at
	$\mathfrak{c}$ corresponding to directions tangent to $M(k,d)^G$ will equal
	the eigenvalues of the Hessian of $\ploss| M(k,d)^G$ at $\mathfrak{c}$.
	Indeed, the Jacobian of $(\grad{\ploss})| M(k,d)^G$ 
	(or $\vectfieldidx{d}$) 
	is equal to the Hessian of $\ploss|M(k,d)^G$.
\end{rem}

Let $S_3$ denote the subgroup of $S_k$ permuting the last three rows of $M(d+2,d)$. Suppose $\mathfrak{c}$ is a type II critical point of $\ploss$. Since $M(d+2,d)^{\Delta S_{d-1}}$ 
is invariant, but not fixed, by the $S_3$-action, it follows by the $S_k \times S_d$-invariance of $\ploss$ that the 
$S_3$-orbit of $\mathfrak{c}$ is a subset of $M(d+2,d)^{\Delta S_{d-1}}$ containing at most six points ($|S_3| = 6$).

We claim that if $\mathfrak{c}$ is fixed
by a non-identity element of $S_3$, then the Hessian of $\mathfrak{c}$ has to be singular. This follows since otherwise at least two of the final three rows of $\mathfrak{c}$ must be parallel and so
$\mathfrak{c}$ lies in a set of critical points of $\cL|M(k,d)^{\Delta S_{d-1}}$ of dimension 
at least one (see \pref{sec:fossil} below). In particular, for a regular family,
(a) critical points cannot be fixed by  a non-identity element of $S_3$, and (b) if $\mathfrak{C} $ is a regular family in  $M(d+2,d)^{\Delta S_{d-1}}$, so is $\sigma \mathfrak{C}$, $\sigma \in S_3$, and 
their paths do not cross. Generally, we regard the six type II families that result from this observation as being essentially the same and focus on just one of them. That choice comes naturally from an
order on the critical points that we discuss shortly. Unlike type I families, the type II family has a rich geometric structure that plays a significant role in the analysis.

The next step is to give an explicit expression for the gradient vector field 
of $\cL$ restricted to $M(d+2,d)^{\Delta S_{d-1}} \approx \real^9$. Given $W 
\in M(d+2,d)^{\Delta S_{d-1}}$, denote the rows of $W$ by 
$\wwi{1},\cdots,\wwi{d+2}$. 
After defining norms, dependent only the row-type, we define angles between 
different rows of $W$ (capital Greek) and angles 
between rows of $W$ and rows of $V$ (lower case Greek). Two subscripts are needed for angles between rows of different row-type, one subscript suffices for the same row type. 
For $p \in [3]$, it is often convenient to set $d_p = d-p$. 

For $a \in [4]$, let $\tau_a$ denote the norm of $\wwi{i} 
\in  M(d+2,d)^{\Delta S_{d-1}}$, where $\wwi{i}$ is of 
row-type 
$a$ and the norm induced from standard Euclidean inner 
product on $M(d+2,d)$. In $\real^N$ coordinates,
$\tau_1 = \sqrt{\xi_1^2 + d_2 \xi_2^2 + \xi_3^2}$, $\tau_2 = \sqrt{d_1\xi_4^2 +  \xi_5^2}$, etc. 

For $i , j \in\is{d-1}$, $ i \ne j$, let $\Theta_1$ denote 
the angle between rows $\wwi{i}$ and $\wwi{j}$ and 
$\theta_1$ denote the angle between row $\wwi{i}$ and 
$\vvi{j}$ (here and below, angles are well-defined 
independently of $i,j$ using symmetry).  For $a,b \in [4]$, 
$a \ne b$, denote the angle between $\wwi{a}$ and $\wwi{b}$ 
by $\Lambda_{ab}$ (that is, the angle between rows of 
row-types $a$ and $b$, $a \ne b$). For $a\in [4]$,
$b \in [2]$, let $\lambda_{ab}$ denote the angle between a $W$-row of type $a$ and a $V$-row of type $b$, where if $a=b\le 2$, we assume $W,V$ are different rows. Note that $\Lambda_{ab}$ is symmetric in $a,b$ but
$\lambda_{ab}$ is not; indeed, $\lambda_{ba}$ is not even defined if $a > 2$. 
Finally, let $\beta_1$ denote the angle between $\wwi{i}$ 
and $\vvi{i}$, where $i \in [d-1]$, and $\beta_2$ denote the 
angle between $\wwi{d}$ and $\vvi{d}$. All these angles may 
be expressed as inverse cosines of expressions in the 
variable $\xi_1,\cdots,\xi_9$. For example
\[
\Lambda_{24} \defoo \cos^{-1}\left(\frac{\langle w^d,w^{d+2}\rangle}{\tau_2\tau_4}\right) = \cos^{-1}\left(\frac{d_1\xi_4\xi_8+\xi_5\xi_9}{\tau_2\tau_4}\right)
\]
The norms and angles are well defined---depend only row-type---on account of symmetry. 

Next we give expressions for the vector field induced on $\real^9$ by  $\nabla \ploss| M(d+2,d)^{\Delta S_{d-1}}$. 
Define
\begin{eqnarray*}
\GGG_1(\tbxi) & = & d_2 \left(\frac{\tau_1\sin(\Theta_1)- \sin(\theta_1)}{\tau_1}\right) + \frac{\big[\sum_{j \ne 4}\tau_j \sin(\Lambda_{1j})\big] -\sin(\lambda_{12})-\sin(\beta_1)}{\tau_1} \\
\GGG_2(\tbxi)& = & d_1\left(\frac{\tau_1\sin(\Lambda_{2 1}) -\sin(\lambda_{21})}{\tau_2}\right) + \frac{\tau_3\sin(\Lambda_{23})+\tau_4\sin(\Lambda_{2 4})  -\sin(\beta_2)}{\tau_2 } \\
\GGG_3(\tbxi) & = & d_1\left(\frac{\tau_1\sin(\Lambda_{3 1}) -\sin(\lambda_{3 1})}{\tau_3}\right) + \frac{\tau_2\sin(\Lambda_{3 2})+\tau_4\sin(\Lambda_{34})-\sin(\lambda_{32})}{\tau_3} \\
\GGG_4(\tbxi) & = &  d_1\left(\frac{\tau_1\sin(\Lambda_{4 1}) -\sin(\lambda_{4 1})}{\tau_4}\right) + \frac{\tau_2\sin(\Lambda_{4 2})+\tau_3\sin(\Lambda_{43})-\sin(\lambda_{4 2})}{\tau_4}  
\end{eqnarray*}
Define nine ``angle'' terms.
\begin{align*}
A_1^1  & = d_2 \Theta_1\xi_2 +  \Lambda_{i2}\xi_4+  \Lambda_{13}\xi_6+\Lambda_{14}\xi_8-\beta_1, \;
A^1_\star  = (\xi_1+ d_3\xi_2) \Theta_1 +  \Lambda_{12}\xi_4+  \Lambda_{13}\xi_6+\Lambda_{14}\xi_8-\theta_1  \\
A^1_2 & =  d_2\Theta_1\xi_3 +\Lambda_{12} \xi_5+\Lambda_{13}\xi_7 + \Lambda_{14} \xi_9-\lambda_{12}\\
A^2_1 & =  (\xi_1 + d_2\xi_2)\Lambda_{1 2} +\Lambda_{23 } \xi_6+ \Lambda_{2 4 } \xi_8 -\lambda_{2 1},\; 
A^2_2  =  d_1\Lambda_{12}\xi_3 +\Lambda_{23} \xi_7+ \Lambda_{24} \xi_9-\beta_2 \\
A^3_1  & =  (\xi_1 + d_2\xi_2)\Lambda_{1 3} +\Lambda_{23} \xi_4 + \Lambda_{34} \xi_8 -\lambda_{3 1}, \;
A^3_2   =  d_1\Lambda_{1 3 }\xi_3 +\Lambda_{23} \xi_5+ \Lambda_{34} \xi_9-\lambda_{32}\\
A^4_1  & =  (\xi_1 + d_2\xi_2)\Lambda_{1 4} +\Lambda_{2 4} \xi_4 + \Lambda_{3 4 } \xi_6 -\lambda_{4 1},\; 
A^4_2   =  d_1\Lambda_{1 4 }\xi_3 +\Lambda_{24} \xi_5+ \Lambda_{34} \xi_7-\lambda_{4 2} 
\end{align*}
Finally, define 
\begin{eqnarray*}
\Omega_1 & = & \pi\left[\xi_1 + d_2\xi_2 + \xi_4+\xi_6 +  \xi_8 -1\right]\\
\Omega_2 & = & \pi\left[d_1\xi_3 + \xi_5+\xi_7 +\xi_9-1\right]
\end{eqnarray*}
Note that $\Omega_1$ is the column sum of any one of the first $d-1$ columns of the matrix $\pi(\Xi(\tbxi) - V)$ and $\Omega_2$ is the sum of column $d$ of $\pi(\Xi(\tbxi) - V)$.

The components 
$(\vectfieldidx{d,1},\cdots,\vectfieldidx{d,9})$ of 
$\vectfieldidx{d}$ are 
given by 

\begin{align*}
\vectfieldidx{d,1}(\bxi) &=\frac{1}{2\pi}(\GGG_1 \xi_1 - 
A_1^1 + 
\Omega_1)  
&\vectfieldidx{d,2}(\bxi) & =\frac{1}{2\pi}(\GGG_1 \xi_2 - 
A_\star^1 
+ 
\Omega_1)  \\
\vectfieldidx{d,3}(\bxi) & =\frac{1}{2\pi}(\GGG_1 \xi_3 - 
A_2^1 + 
\Omega_2) &&\\
\vectfieldidx{d,4}(\bxi) &= \frac{1}{2\pi}(\GGG_2 \xi_4 - 
A_2^1 + 
\Omega_1) 
&\vectfieldidx{d,5}(\bxi) & =\frac{1}{2\pi}(\GGG_2 \xi_5 - 
A_2^2 + 
\Omega_2) \\
\vectfieldidx{d,6}(\bxi) & =\frac{1}{2\pi}(\GGG_3 \xi_6 - 
A^3_1 + 
\Omega_1) 
&\vectfieldidx{d,7}(\bxi) & =\frac{1}{2\pi}(\GGG_3 \xi_7 - 
A^3_2 + 
\Omega_2) \\
\vectfieldidx{d,8}(\bxi) & =\frac{1}{2\pi}(\GGG_4 \xi_8 - 
A^4_1 + 
\Omega_1) 
&\vectfieldidx{d,9}(\bxi) & =\frac{1}{2\pi}(\GGG_4 \xi_9 - 
A^4_2 + 
\Omega_2)
\end{align*}
and so the critical point equations on $\real^9$ are 
$\vectfieldidx{d,i}(\bxi) = 0$, $i\in[9]$. That is, 
\begin{align*}
\frac{1}{2\pi}(\GGG_1 \xi_1 - A_1^1 + \Omega_1)&=0  
&\frac{1}{2\pi}(\GGG_1 \xi_2 - A_\star^1 + \Omega_1)&=0  \\
\frac{1}{2\pi}(\GGG_1 \xi_3 - A_2^1 + \Omega_2)&=0 &&\\
\frac{1}{2\pi}(\GGG_2 \xi_4 - A_2^1 + \Omega_1)&=0 
&\frac{1}{2\pi}(\GGG_2 \xi_5 - A_2^2 + \Omega_2)&=0 \\
\frac{1}{2\pi}(\GGG_3 \xi_6 - A^3_1 + \Omega_1)&=0 
&\frac{1}{2\pi}(\GGG_3 \xi_7 - A^3_2 + \Omega_2)&=0 \\
\frac{1}{2\pi}(\GGG_4 \xi_8 - A^4_1 + \Omega_1)&=0 
&\frac{1}{2\pi}(\GGG_4 \xi_9 - A^4_2 + \Omega_2)&=0
\end{align*}

If $\tbxi \in \real^9$ determines a type II critical point, 
then $\xi_1 \arr +1$ as $d \arr \infty$. Numerics also 
indicate that for type II one of $\xi_5,\xi_7, \xi_9$ 
converges to $-1$ as $d \arr \infty$. 
Permuting with an element of $S_3$, we may and shall hypothesize that 
$\xi_{9} \arr -1$ as $d \arr \infty$.  It helps to use some 
geometry concerning the rows $\wwi{d}, \wwi{d+1}$ of $W = 
\Xi(\tbxi)$. Let $\uui{d} = \sum_{i\in[d-1]} \vvi{i}\in 
\real^d$ and $\mathbb{F} \subset \real^d$ be the 
2-dimensional subspace spanned by
$\vvi{d}, \uui{d}$. Observe that $\wwi{d}, \wwi{d+1}, 
\wwi{k}, \vvi{d} \in \mathbb{F}$ and so are coplanar. By the 
analyticity properties of regular families,  
$\wwi{d}$ and $\wwi{d+1}$ cannot be parallel.
Hence either $\beta_2 < \lambda_{32}$ or $\beta_2 > \lambda_{32}$: \emph{curves of regular families do not cross}. In the first case $\Lambda_{23} = \lambda_{32} - \beta_2 $; in 
the second $\Lambda_{23} =  \beta_2 - \lambda_{32}$. Composing with a unique $\sigma \in S_3$, fixing the last row, we may always assume $\Lambda_{23} = \lambda_{32} - \beta_2$.  Numerics indicate that
$\lambda_{32} > \beta_2 > \pi/2 > \Lambda_{23} > 0$---but this is not assumed in what follows.

The idea now is to take formal FPS expansions for the components of a type II 
critical point, substitute in the critical point equations described above, 
equate like coefficients and thereby obtain FPS solutions. Guided by the 
numerics, we seek a power series in $d^{-\frac{1}{2}}$. Granted our hypothesis 
on the constant terms in the FPS, knowledge of vanishing coefficients (see 
\pref{sec:typeIorII}), and an easy computation comparing like terms giving 
$c_{3,2}$, we have
\begin{align*}
c_{1,0}& = 1,\; c_{1,i} = 0,\; i \in [2], \;\;c_{2,i} = 0,\; i \le 3, \;\; 
c_{3,i} = 0,\; i < 2,\; c_{3,2} = 2, \\
c_{4,i}& = 0,\; i < 2, \;\;c_{5,0} = 0, \;\;c_{6,i} = 0, i < 2,\;\; c_{7,0} = 
0,\;\; c_{8,i} = 0,\; i < 2,  \\
c_{9,0} & = -1.
\end{align*}
For notational 
clarity, we relabel the 9 unknown coefficients $c_{i,j}$ giving the next terms 
in the FPS so that we aim to find $c_3,e_4,f_3,g_2,h_1,p_2, q_1,a_2, b_1 
\in\real$ such that
\begin{align*}
\xi_1 & =1 + c_3 d^{-\frac{3}{2}} + \cdots  \;\;
\xi_2  = e_4d^{-2} + \cdots  \;\;
\xi_3  = 2 d^{-1} + f_3 d^{-\frac{3}{2}}\cdots\\
\xi_4 & = g_2 d^{-1} + \cdots \;\qquad
\xi_5  = h_1 d^{-\frac{1}{2}} + \cdots \\
\xi_6 & = p_2 d^{-1} + \cdots \;\qquad
\xi_7  = q_1 d^{-\frac{1}{2}} + \cdots\\
\xi_8 & = a_2 d^{-1} + \cdots  \;\qquad
\xi_9  = -1 + b_1d^{-\frac{1}{2}} + \cdots 
\end{align*}
The condition we gave on angles holds if and only if $h_1 p_2 > g_2 q_1$ (both sides are negative). 
Set $R_2 = \sqrt{g_2^2 +h_1^2}$, $R_3 = \sqrt{p_2^2 + q_1^2}$. We derive expressions for the angles $\Lambda_{ab}$, $a,b \in \{2,3,4\}$ and find that
\[
\Lambda_{34} = \Lambda_{34}^0 + O(d^{-\frac{1}{2}}),\;\;\Lambda_{23 } = \Lambda_{23}^0 + O(d^{-\frac{1}{2}}),\;\;\Lambda_{24} = \Lambda_{24}^0 + O(d^{-\frac{1}{2}}),
\]
where  
\begin{align*}
 \Lambda_{24}^0&=  \sin^{-1}(g_2/R_2)\in (0,\pi/2)\\
\Lambda^0_{23}&= \sin^{-1}((h_1p_2- g_2q_1)/(R_2 R_3)) \in (0,\pi/2) \;(\text{since } h_1p_2 > g_2 q_1)\\
\Lambda_{34}^0&= \sin^{-1}(p_2/R_3)\in (0,\pi/2)
\end{align*}
Using standard trigonometric formulas, we deduce the relationship  $\Lambda_{23} + \Lambda_{34} = \Lambda_{24}$ and, letting $d \arr \infty$, $\Lambda^0_{23} + \Lambda^0_{34} = \Lambda^0_{24}$.
We have similar expressions for $\beta_2$ and $\lambda_{a2}$, $a \in  \{3,4\}$:
\[ 
\beta_2= \beta_2^0 + O(d^{-\frac{1}{2}}) \;\;  \lambda_{3 2}= \lambda^0_{32} + O(d^{-\frac{1}{2}}) \;\;\lambda_{42} = \lambda^0_{42} + O(d^{-\frac{1}{2}})
\]
\[
\beta^0_2= \cos^{-1}(h_1/R_2) \in (\pi/2,\pi) \;\; \lambda^0_{32}=  \cos^{-1}(q_1/R_3) \in (\pi/2,\pi)\;\;  \lambda_{42}^0 = \pi 
\]
It follows that $\Lambda_{24} + \beta_2  = \lambda_{42}$ and $\Lambda_{34} + \lambda_{32} = \lambda_{42}$, with the same identities holding between the constant terms by letting $d \arr \infty$. 
In particular, all the constant terms for the FPS expansions of these angles can be expressed in terms of $R_2, \Lambda_{24}^0$ and $R_3, \Lambda_{34}^0$ (polar coordinates on $(g_2,h_1)$- and $(p_2,q_1)$-space).

Equating like coefficients of terms in the equations and noting in particular that the coefficients $\pi(e_4+g_2 + p_2 +a_2)$ of $d^{-1}$ in $\Omega_1$, and $\pi(f_3+h_1 + p_1 +b_1)$ of $d^{-\frac{1}{2}}$ in
$\Omega_2$ are zero, we derive a system of nine nonlinear equations that determine $c_3,e_4, f_3, g_2, h_1, p_2,q_1,a_2, b_1$:
\begin{align*}
 c_3 -b_1 + R_2 + R_3& = 0\\
 e_4 + g_2 + p_2 + a_2& =0 \\
 f_3 + h_1 + q_1 + b_1& = 0\\
 g_2\left(\frac{c_3R_2^2-2g_2h_1}{R_2^3}+\frac{-b_1g_2+h_1a_2-g_2q_1 + h_1p_2}{R_2^2 }\right)& =  \Lambda_{23}^0p_2 + \Lambda_{24}^0a_2+\frac{\pi}{2}e_4 -\frac{2h_1}{R_2}  \\
 h_1\left(\frac{c_3R_2^2-2g_2h_1}{R_2^3}+\frac{-b_1g_2+h_1a_2-g_2q_1 + h_1p_2}{R_2^2 }\right)& =  \Lambda_{23}^0q_1 + \Lambda_{24}^0b_1+\frac{\pi}{2}f_3 -\frac{2g_2}{R_2}+a_2  \\
 p_2\left(\frac{c_3R_3^2-2p_2q_1}{R_3^3}+\frac{-b_1p_2+q_1a_2 + p_2h_1 - q_1g_2}{R_3^2 }\right)& =  \Lambda_{23}^0g_2 + \Lambda_{24}^0a_2+\frac{\pi}{2}e_4 -\frac{2q_1}{R_3}  \\
 q_1\left(\frac{c_3R_3^2-2p_2q_1}{R_3^3}+\frac{-b_1p_2+q_1a_2 + p_2h_1 - q_1g_2}{R_3^2 }\right)& =  \Lambda_{23}^0h_1 + \Lambda_{24}^0b_1+\frac{\pi}{2}f_3 -\frac{2p_2}{R_3}+a_2  \\
2 - \frac{\pi}{2}e_4  & =  (\pi-\Lambda^0_{24}) g_2 + (\pi - \Lambda^0_{34}) p_2 + \pi a_2 \\
 c_3- \frac{\pi}{2}f_3+g_2 + p_2 & = (\pi-\Lambda^0_{24})h_1 + (\pi - \Lambda^0_{34})q_1 + \pi b_1
\end{align*}
It is possible to reduce to a system of four equations in $R_2, \Lambda^0_{24}, R_3,\Lambda^0_{34}$ (or two in $\Lambda^0_{24},\Lambda^0_{34}$), solve and then express the coefficients $c_3,e_4, \cdots, b_1$ in terms of 
$R_2, \Lambda^0_{\ell k}, R_3,\Lambda^0_{dk}$. In practice, we either use 
Newton-Raphson method on the original system, using an initialization based on 
numerics, or
eliminate $c_3,e_4,f_3$ from the 9 equations and solve the resulting nonlinear system of 6 equations using Newton-Raphson. We used the reduction to a system of six equations, and found that 
\[
\begin{matrix}
c_3& =&\!\! -0.5748287640041448964\ldots&  e_4& = &\;\!\! -1.6165352425422284608\ldots \\
f_3& =&  \;0.2969965493462016520\ldots & && \\
g_2& =&  \; 0.7877659431796313120\ldots&  h_1& = &\! -1.1161365378487412475\ldots & \\
 p_2& =& \; 0.1562694812799615923\ldots& q_1& = &-0.4248280138040598900\ldots   & \\
a_2& =&\,  0.6724998180826355564\ldots& b_1& = & \;\;\, 1.2439680023065994855\ldots &
\end{matrix}
\]
Newton's method was initialized using numerical data for the critical points 
when $d=10^{4}$ to get rough estimates for the coefficients; the original 
computation was done in long double precision. The values were compared with 
those obtained by directly solving FPS equations corresponding to different 
orders of the FPS terms (see \pref{sec:typeIorII}), and through a high 
precision computation of 
the critical point for $d = 10^{512}$---the values of the components of the 
gradient at the critical 
point were all less than $10^{-4000}$ and matched with those obtained by 
solutions of the equations above. Note that
for $d = 10^{512}$, one expects to be able to read off the required 
coefficients from the computed critical points to 250 or more decimal places of 
accuracy (the series is in integer powers of $d^{-\frac{1}{2}}$).
\begin{rem}\label{rem:sys9}
It is straightforward to check that the Jacobian of the 9 variable system in $(c_3,\cdots,b_1)$ with respect to $(c_3,\cdots,b_1)$ is non-singular (for this, we need expressions for the angles $\Lambda^0_{ab}$, 
$a,b \in \{2,3,4\}$, $a \ne b$, in terms of $(c_3,\cdots,b_1)$).   This is significant for the next and final step.
\end{rem}

Having determined $c_3,e_4, \cdots, b_1$, we set $t^{-\frac{1}{2}} = s$ and 
define $\widetilde{\txi}_i(s)$, $ i \in [9]$ by 
\begin{align*}
\xi_1(s)& = -1 + s^3 \widetilde{\txi}_1(s),\; \xi_2(s) = s^4 \widetilde{\txi}_2(s),\; \xi_1(s) = 2 s^{-1} + s^3\\
\xi_4(s) & = s^2 \widetilde{\txi}_4(s),\; \xi_5(s) = s \widetilde{\txi}_5(s)\\
\xi_6(s) & = s^2 \widetilde{\txi}_6(s),\; \xi_7(s) = s \widetilde{\txi}_5(s)\\
\xi_8(s) & = s^2 \widetilde{\txi}_8(s),\; \xi_9(s) = -1 + s \widetilde{\txi}_9(s)
\end{align*}
where the values of $\widetilde{\txi}_1(0),\ldots,\widetilde{\txi}_9(0)$ are given by $(c_3,\cdots,b_1)$ respectively. 
Substitute in the critical point equations---with $d$ everywhere replaced by 
$s^{-2}$---and cancel the powers of $s$ that occur in each equation.
The Jacobian of this system with respect to $(\widetilde{\txi}_1,\ldots,\widetilde{\txi}_9)$ at $s = 0$ is non-singular (this uses \pref{rem:sys9}). 
Finally, use the real analytic version of the implicit function theorem to obtain the FPS.

\begin{rem}\label{rem:notes}
Although the argument only gives the convergence of the FPS for sufficiently 
large $d$ (that is, sufficiently small $d^{-\frac{1}{2}}$), it appears that 
the series converges for small $d$, possibly all $d$ for which the problem is 
defined. This is similar to what happens when $k = d$ 
\cite{ArjevaniField2020}.\\
\end{rem}
\subsection{Type II, $k = d+1$}
If $k = d+1$, we can reduce the analysis to solving a single equation $p(\vartheta) = 0$, where $p$ is a polynomial in $\vartheta, \sin(\vartheta)$ and $\cos(\vartheta)$. We solve $p = 0$ directly using Newton's method, initializing with 
a value of $\vartheta$ suggested by numerics. For completeness, we give $p$ explicitly as well as the coefficients of interest in the associated FPS. 

We have $\text{dim}(M(d+1,d)^{\Delta S_{d-1}}) = 7$. The critical point equations are read off easily from those we gave for $k = d+2$: drop the last two equations and all terms that involve
the variables $\xi_8,\xi_9$ or an angle indexed with a `4'. Just as in the case when when $k = d+2$, we reduce to finding
the initial coefficients $c_3,e_4,f_3,g_2,h_1,p_2,q_1$ in the FPS expansion
\begin{align*}
\xi_1 & = 1 + c_3 d^{-\frac{3}{2}} + \cdots, \;
\xi_2  = e_4d^{-2} + \cdots,  \;
\xi_3  = 2 d^{-1} + f_3d^{-\frac{3}{2}} + \cdots,\\
\xi_4 & = g_2 d^{-1} + \cdots,\qquad\
\xi_5  = h_1 d^{-\frac{1}{2}} + \cdots \\
\xi_6 & = p_2 d^{-1} + \cdots, \;\qquad
\xi_7  = -1 + q_1d^{-\frac{1}{2}} + \cdots 
\end{align*}
As in the case $k = d+2$, we obtain a system of seven nonlinear equations.
\begin{align*}
 c_3 -b_1 + R_2& = 0\\
 e_4 + g_2 + a_2& =0 \\
 f_3 + h_1 + b_1& = 0\\
 g_2\left(\frac{c_3R_2^2-2g_2h_1}{R_2^{\frac{3}{2}}}+\frac{-b_1g_2+h_1a_2}{R_2^2}\right)& =  \vartheta a_2 +\frac{\pi}{2}e_4 -\frac{2h_1}{R_2}  \\
 h_1\left(\frac{c_3R_2^2-2g_2h_1}{R_2^{\frac{3}{2}}}+\frac{-b_1g_2+h_1a_2}{R_2^2}\right)& = \frac{\pi}{2}f_3 - \frac{2g_2}{R_2} +  \vartheta b_1 + a_2 \\
 \frac{\pi}{2}e_4 + 2 + \vartheta g_2& = 0\\
 c_3+ \frac{\pi}{2}f_3+g_2 + \vartheta h_1& =0, 
\end{align*}
where $R_2 = \sqrt{g_2^2 + h_1^2}$ and $\vartheta = \Lambda_{23}^0 = \sin^{-1}\left(\frac{g_2}{R_2}\right) \in (0,\frac{\pi}{2})$.

We may reduce to a single equation $p(\vartheta) = 0$, where 
$p = AQ - BP$ and
\begin{eqnarray*}
A(\vartheta)&=&\frac{2}{2-\pi}\left[\frac{\sin(2\vartheta)}{2}\big(1-\sin(\vartheta)\big)\big(\vartheta - \frac{\pi}{2}\big) + \sin(\vartheta)(1-\sin(\vartheta))^2\right] \\
&& +\sin(\theta)\left[2\theta - \frac{2\theta^2}{\pi}-1 - \frac{\sin(2\theta)}{2}\big(\frac{2\theta}{\pi}-1\big)\right] \\
B(\vartheta)&=& \frac{2}{2-\pi} \left[-\cos(\vartheta)\big(\frac{\pi}{2}-\vartheta\big)^2 + \big(\frac{\pi}{2}-\vartheta\big)(1-\sin(\vartheta))(2-\sin^2\vartheta)-\cos(\vartheta)(1-\sin(\vartheta))^2\right]\\
&& + \left[\cos(\vartheta)\big(1-\frac{\pi}{2}\big) - \sin^3\vartheta \big(\frac{2\vartheta}{\pi}-1\big)\right]\\
P(\vartheta) & = & 2 - \frac{4\vartheta}{\pi} - \frac{2\sin(2\vartheta)}{\pi}-2\cos^3\vartheta \\
Q(\vartheta) & = & 2 \sin^3\vartheta -\frac{4}{\pi}\sin^2\vartheta
\end{eqnarray*}
Using Newton-Raphson, the required solution to $p(\vartheta) = 0$ is given 
by
\[
\vartheta = 0.58416413506022510436594641534260755532740719514252671834097577387592202\ldots
\] 
with $|p'(\vartheta)| > 0.4$ and $|p(\vartheta)| < 10^{-2400}$

We have $g_2 = R_2\sin(\vartheta)$, $h_1 = -R_2\cos(\vartheta)$ and $R_2 = -P(\vartheta)/A(\vartheta)$. It follows easily that  
the remaining coefficients $c_3,e_2,\cdots,b_1$ are uniquely determined by $\vartheta$.  We find that 
\[
\begin{matrix}
c_3 =&\;\;\, -0.57228787893585490607\ldots&  e_4 = &\;\;\, -1.61458989052095508224\ldots \\
f_3 = & \;0.29629854644604431015\ldots & & \\
g_2 = & \; 0.91787878976036618322\ldots&  h_1 = & -1.38833511087258399162\ldots & \\
a_2 = &\; 0.69671110076058889902\ldots& b_1 = & \;\;\, 1.0920365644265396815\ldots &
\end{matrix}
\]
The existence of the FPS now follows the method for $k = d+2$.
\subsection{Type I, $k = d+1, d+2$}

We conclude with a brief discussion of the type I family 
when $k = d+1, d+2$. This behaves rather differently from 
type II. For example, no row of a critical point converges 
to $\vvi{d}$ 
as $d \arr \infty$. Instead, one row converges to
$\vvi{d}/2$, another to $-\vvi{d}/2$. Moreover, the 
fractional power series are in $d^{-\frac{1}{4}}$ rather 
than $d^{-\frac{1}{2}}$. As a result, the convergence of 
critical points as $d \arr \infty$ is slow. 
If $k = d+1$, explicit expressions for the critical coefficients are given in 
Example 1 of \pref{sec:family} (see \pref{sec:typeIorII} for a detailed 
derivation). The analysis then proceeds as in the type II case. 

If $k = d+2$, we again have the constant coefficients  $1/2, -1/2$ for row-types $2,3$ respectively (if necessary after a permutation by an element of $S_3$.
Here is probably easiest (certainly faster), to obtain the system of nine equations, as we did for type II, and then solve using Newton-Raphson. In brief, after some work the critical coefficients 
for the FPS in $d^{-\frac{1}{4}}$ are
given by 
\[
\begin{matrix}
c_6& =& \,2.6472714633048307498\ldots&  e_7& = &\!\!\!\! -1.0684533932698202809\ldots \\
f_5& =&\!\!  -0.8644915139550179823\ldots & && \\
g_3& =&  \; 0.5342266966349101404\ldots&  h_1& = &\!      0.4322457569775089911\ldots & \\
p_3& =& \;  0.5342266966349101404\ldots& q_1& = &         0.4322457569775088806\ldots   & \\
a_4& =&\,   1.2534701553854549462\ldots& b_1& = & \hspace*{-0.1in} -0.8753051450722888701\ldots &
\end{matrix}
\]
Note that $c_4 = e_4 = 2$,  $f_4 = 1$, $g_3 = p_3$ and $h_1 = q_1$, just as for $k = d+1$.
The existence of the FPS follows as above. 

\subsection{Hessian spectrum for type I}
In Table 2 we give the Hessian spectrum associated to the 
standard and trivial representations for type I points when $k = d+1, d+2$  for 
$d = 10, 100$.

\begin{table}[h]\label{table:typeIevs}
        \begin{center}
                \begin{tabular}{ |c||c|c||c|c| } \hline
                        Isotypic comp.   & $k=d+1$ & $k=d+2$ & $k=d+1$ & $k=d+2$ \\\hline
                        $d$ &$ 10$ & $10$ & $100$ & $100$\\\hline\hline
                          & $0.01859,0.03574$ &$0.00613, 0.01715$& $0.006647,0.05914$ &$0.006467,0.01348$   \\
            $\mathfrak{t}$ &$0.08472,0.25925$&$0.04434, 0.05308$& $0.20056, 0.27534$  &$ 0.06335,0.09132$ \\
                        &$1.5634,2.90219$&$0.2309, 0.3080$& $15.746, 25.4661$ &$ 0.2697,0.4127$ \\
                                  &$3.3678$&$1.6453,3.2022$  &$26.055$&$15.846,25.74$\\
                        &&$3.7551$  & &$26.395$\\

                        \hline
                        &$-0.00230, 0.04343$  &$-0.03903,0.00423$  &$0.03178,0.0680$  &$-0.03432,0.03707$\\
        $\mathfrak{s}_{d-1}$ &$0.1135,0.2324$   &$0.04824,0.1206$  &$0.09210,0.24363$ &$0.07250,0.09303$\\
                             &$0.3915,2.9398$&$ 0.2630, 0.4559$  &$0.48426,25.478$ &$0.3103,0.5329$\\
                         &  &$3.2410$ &&$25.751$\\
                        \hline
                \end{tabular}
        \end{center}
        \caption{Type I critical points with isotropy $\Delta S_{d-1}$ \& the
        Hessian eigenvalues associated to the trivial and the
                standard representation to four decimal places when 
                $k = d+1, d+2$ and 
               $d = 10$ and $100$. The spectrum associated to the 
               $\mathfrak{x}$- and $\mathfrak{y}$-representations is strictly 
               positive and not shown.       }
\end{table}
Referring to the table, the addition of one extra neuron results in the type I critical point becoming a saddle when $d = 10$ (it defines a spurious minimum if $k = d = 10$) but is a spurious 
minimum for $d = 100$. If we add a second neuron, the type I critical point becomes a saddle for $d = 100$; most likely a saddle for all $d \ge 4$ (certainly sufficiently large $d$ by our results).

\subsection{Type of regular families and derivation of case (3) in \pref{ex: 
fpsapp}}\label{sec:typeIorII}

Evaluating formal Puiseux series at critical points gives rise to 
algebraic relations between the Puiseux series coefficients, allowing 
one to argue about the structure of regular families 
(a-priori, independently of their existence). We show how these 
relations can be used to deduce that for $k=d+1$, any regular family of $\Delta 
(S_{d-1}\times S_1)$-critical points with $\base = 4$ must be either 
type I, type II or have all its initial terms vanish. Other isotropy 
types and pairs of $d$ and $k$ are addressed similarly. Although only 
the diagonal of the main $(d-1)\times(d-1)$-block is required for 
determining the type of a family (see \pref{def: types}), we 
evaluate all the entries which belong to the $(d-1)\times d$-upper 
block to low-order terms. This makes the derivation of the 
eigenvalue expressions given in (\pref{sec:xy_evs_pf}) more 
transparent, and further serves as a preparation 
for the detailed derivation below for the type I $\Delta 
(S_{d-1}\times S_1)$-critical points given in \pref{ex: fpsapp}.

As in the previous sections, denote the FPS coefficients corresponding to 
$\xi_1,\xi_2,\dots,\xi_7$ by $c_i, e_i, f_i, g_i, h_i, p_i, q_i$, and let 
$\partder{i}$ denote the $i$'th component of the vector field 
$\vectfieldidx{}$. Note that regularity assumptions imply that 
all coefficients with (strictly) negative index must vanish (e.g., necessarily, 
$c_{-1}=0$). We show that for any regular family of critical points in 
$M(d+1,d)^{\Delta(S_{d-1}\times S_1)}$ with $\base=4$, it follows 
that $c_0 = \pm1$ and $c_1 = e_0 = e_1 = e_2 = e_3 = f_0 = f_1 = f_2 
= f_3 = 0$. The following notation comes handy when handling 
expressions involving Puiseux series coefficients. Given a 
 Puiseux series $E = \sum_{j=j_0}^\infty \eta_j d^{-\frac{j}{4}}$ where 
$j_0\in\mathbb{Z}$, let $[d^{-\frac{j}{4}}]E \defeq \eta_j$.

Observe that 
\begin{align*}
	[d] \partder{1} = \dfrac{e_0}{2}\quad \text{ and }\quad [d] 
	\partder{3} = \dfrac{f_0}{2}.
\end{align*}
Since the gradient entries vanish at critical points, necessarily, $e_0=f_0=0$. 
Similarly,
\begin{align*}
	[d^{\frac{3}{4}}] \partder{1} = \dfrac{e_1}{2}\quad \text{ and }\quad 
	[d^{\frac{3}{4}}] 
	\partder{3} = \dfrac{f_1}{2},
\end{align*}
imply $e_1 = f_1 = 0$. Assuming momentarily $c_0\neq0$, we have
\begin{align*}
	&[d] \partder{1} = \frac{c_{0} \sqrt{- 
	\frac{e_{2}^{4}}{c_{0}^{4} + 2 
	c_{0}^{2} e_{2}^{2} + e_{2}^{4}} + 1}}{2 \pi} - \frac{c_{0}}{2 \pi 
	\sqrt{c_{0}^{2} + e_{2}^{2}}},\\
	&[d^{\frac{1}{2}}] \partder{2} = \frac{e_{2} \sqrt{- 
	\frac{e_{2}^{4}}{c_{0}^{4} + 2 
	c_{0}^{2} e_{2}^{2} + e_{2}^{4}} + 1}}{2 \pi} - \frac{e_{2} 
	\operatorname{acos}{\left(\frac{e_{2}^{2}}{c_{0}^{2} + e_{2}^{2}} 
	\right)}}{2 \pi} + \frac{e_{2}}{2} - \frac{e_{2}}{2 \pi \sqrt{c_{0}^{2} + 
	e_{2}^{2}}}.
\end{align*}
Dividing the first expression by $c_0$ and using the resulting expression to 
simplify the second one gives
\begin{align*}
\frac{e_{2}}{2}- \frac{e_{2} 
\operatorname{acos}{\left(\frac{e_{2}^{2}}{c_{0}^{2} + e_{2}^{2}} 
\right)}}{2 \pi}  =0 .
\end{align*}
If, by way of contradiction, $e_2$ is assumed to be non-zero then 
the preceding equation gives
\begin{align*}
\operatorname{acos}{\left(\frac{e_{2}^{2}}{c_{0}^{2} + e_{2}^{2}} 
\right)} = \pi,
\end{align*}
a contradiction, as the argument of $\arccos$, 
$\frac{e_{2}^{2}}{c_{0}^{2} + e_{2}^{2}}$, is nonnegative. Hence, $e_2=0$ and 
\begin{align*}
	[d] \partder{1} &= c_0\prn*{\frac{1}{2\pi} - \frac{1}{2\pi|c_0|}},
\end{align*}
whence $c_0 = \pm1$. Evaluating $[d^{\frac{3}{4}}] \partder{1}$, 
$[d^{\frac{1}{2}}] \partder{3}$, $[d^{\frac{1}{4}}] \partder{2}$ and  
$[d^{\frac{1}{4}}] \partder{3}$ then yields $c_1 = e_3 = f_2 = f_3 = 0$, as 
required.

The case $c_0=0$ is addressed similarly. We shall only point out a 
possible course of derivation rather than provide the full expressions. Recall 
that $e_0 = e_1 = f_0 = f_1 =0$ holds regardless of the value assigned to 
$c_0$. Evaluating $[d^{\frac{1}{2}}] \partder{2}$ and $[d^{\frac{1}{2}}] 
\partder{3}$ gives $e_2 = f_2= 0$. Evaluating $[d^{\frac{1}{4}}] \partder{2}$ 
gives $e_3 = 0$. By 
$[d^{\frac{1}{2}}] \partder{4}$ and $[d^{\frac{1}{2}}] \partder{6}$, 
$g_0=p_0=0$. Lastly, evaluating $[d] \partder{5}$ and $[d] 
\partder{7}$ gives $h_0=q_0=0$, concluding the derivation.

The procedure just presented is based on the direct approach described in 
\cite[Section 8]{ArjevaniField2020} by which one extracts coefficients by 
directly solving the FPS equations, exactly or numerically, to 
an increasing order. One proceeds until sufficient information has been 
obtained so as to establish the existence of an FPS and estimate the 
Hessian 
spectrum to a desired order. Below we give a detailed derivation of the 
type I 
$\Delta (S_{d-1}\times S_1)$-minima in~\pref{ex: fpsapp} to demonstrate 
how the approach may be used in~practice.

\paragraph{A detailed derivation of case (3) in \pref{ex: 
fpsapp}.} (Notation and assumption as above.)
For any family of $\Delta (S_{d-1} \times 
S_1)$-critical points, $c_0 = \pm1$ and $c_1 = e_0 = e_1 = e_2 = e_3 = f_0 = 
f_1 = f_2 = f_3 = 0$. The family of critical points given in \pref{ex: fpsapp} 
is type I, and so $c_0=-1$. The derivation proceeds as follows.

\begin{enumerate}[leftmargin=*]
	\item Observe that
	\begin{align*}
		[d^0] \partder{2} &= \frac{e_{4}}{4} + \frac{g_{0}}{4} + \frac{p_0}{4} 
		- 
		\frac{1}{2},\\
		[d^0] \partder{3} &= \frac{f_{4}}{4} + \frac{h_{0}}{4} + \frac{q_0}{4} 
		- 
		\frac{1}{4}.
	\end{align*}
	We use these relations to substitute $e_4$ and $f_4$ for lower-order terms. 
	That is,
	\begin{align*}
		e_{4} = - g_{0} - p_0 + 2,\\
		f_{4} = - h_{0} - q_0 + 1.
	\end{align*}
	
	\item By $[d^{\frac{1}{2}}]\partder{1} = 
	\frac{c_{2}}{2 \pi} - 
	\frac{\abs{g_{0}}}{2 
		\pi} - 
	\frac{\abs{p_{0}}}{2 \pi}$, $c_2 = \abs{g_0} + 
	\abs{p_0}$.
	\item \label{item:cases} We now show that $g_0=p_0=0$. If $g_0=0$ (resp. 
	$p_0=0$) then by $[d^0]\partder{6} = p_0\prn*{\frac{1}{4}-\frac{1}{2\pi}}$
	(resp. $[d^0]\partder{4} = q_0\prn*{\frac{1}{4}-\frac{1}{2\pi}}$), $p_0=0$ 
	(resp. $g_0=0$). Assume then, by way of contradiction, that both $g_0$ and 
	$p_0$ are non-zero, i.e., $g_0\neq0$ and $p_0\neq0$. Then $g_0$ 	and 
	$p_0$ 
	must satisfy the following four equations (effectively two, by 
	symmetry) obtained by evaluating $[d^0]\partder{4}, [d^0]\partder{5}, 
	[d^0]\partder{6}$ and $[d^0]\partder{7}$:
	\begin{align}
		\quad  - \frac{g_{0}}{2 \pi} + \frac{g_{0}}{4} - \frac{g_{0} 
			\abs{p_0}}{2 \pi \abs{g_{0}}} - \frac{p_0 
			\operatorname{acos}{\left(\frac{g_{0} p_0}{\abs{g_{0}} 
			\abs{p_0}} 
				\right)}}{2 \pi} + \frac{p_0}{4}&=0,\\
	 - \frac{h_{0}}{2 \pi} + \frac{h_{0}}{4} - \frac{h_{0} 
			\abs{p_0}}{2 \pi \abs{g_{0}}} - \frac{q_0 
			\operatorname{acos}{\left(\frac{g_{0} p_0}{\abs{g_{0}} 
			\abs{p_0}} 
				\right)}}{2 \pi} + \frac{q_0}{4} &=0,\\	
		\quad  - \frac{g_{0} \operatorname{acos}{\left(\frac{g_{0} 
					p_0}{\abs{g_{0}} \abs{p_0}} \right)}}{2 \pi} 
					+ \frac{g_{0}}{4} - 
		\frac{p_0 \abs{g_{0}}}{2 \pi \abs{p_0}} - \frac{p_0}{2 
		\pi} + 
		\frac{p_0}{4} &=0,\\
		 - \frac{h_{0} \operatorname{acos}{\left(\frac{g_{0} 
					p_0}{\abs{g_{0}} \abs{p_0}} \right)}}{2 \pi} 
					+ \frac{h_{0}}{4} - 
		\frac{q_0 \abs{g_{0}}}{2 \pi \abs{p_0}} - \frac{q_0}{2 
		\pi} + 
		\frac{q_0}{4} &=0.
	\end{align}
	If $g_0,p_0>0$ or $g_0,p_0<0$, then by
	\begin{align*}
		[d^0] \partder{4} = - \frac{g_{0}}{2 \pi} + \frac{g_{0}}{4} - 
		\frac{p_0}{2 
		\pi} + 
		\frac{p_0}{4},
	\end{align*}
	it follows that $g_0 = -p_0$, a contradiction. If $p_0<0<g_0$ then by 
	$[d^0] \partder{4}$	again, $p_0=g_0$, still a contradiction. The case 
	where $g_0<0<p_0$ is 
	treated similarly (and in fact follows by symmetry). Thus, necessarily, 
	$g_0=p_0=0$, and so $c_2=0$ and $e_4 = 2$.
	
	\item \label{typeI:c3} Next, we have 
	$[d^{\frac{1}{4}}]\partder{1} = \frac{c_{3}}{2 
	\pi} 
	- 
	\frac{\abs{g_{1}}}{2 
		\pi} - \frac{\abs{p_1}}{2 \pi}$ and 
		$[d^{-\frac{1}{4}}]\partder{2} = 
		\frac{e_5}{4}+
	\frac{g_1}{4}+\frac{p_1}{4}$. Solving for $c_3$ and $e_5$ shows, by the 
	same argument used in (\ref{item:cases}), that any of the four cases 
	concerning the (strict) signs of $g_1$ and $p_1$ yields a contradiction. 
	Thus, $g_1=p_1=0$, and so $c_3=e_5=0$.

	\item We have $[d^{0}] \partder{1} = \frac{c_{4}}{2 \pi} - 
	\frac{\sqrt{g_{2}^{2} + 
			h_{0}^{2}}}{2 \pi} - \frac{\sqrt{p_2^{2} + q_0^{2}}}{2 \pi} - 
			\frac{1}{2 	\pi}$. In addition, by 
			$[d^{-\frac{1}{4}}]\partder{3}$ we have 
			$f_5=-h_1-q_1$, 
		and	by $[d^{-\frac{1}{2}}]\partder{2}, e_6 = 
		-g_2-p_2$.
	\item  Using $c_4 = \sqrt{g_2^2 + h_0^2} + \sqrt{p_2^2 + q_0^2} + 1$, we 
	obtain 
	two (effectively one by symmetry) equations corresponding 
	respectively  to $[d^0]\partder{5}$ and $[d^0]\partder{7}$:
	\begin{align*}
		\text{(I)}\quad 0 &= - \frac{h_{0}}{2 \pi} + \frac{h_{0}}{4} - 
		\frac{h_{0} 
			|g_2|}{2 \pi 
			\left(g_{2}^{2} + h_{0}^{2}\right)} + \frac{h_{0} \sqrt{\left(g_{2} 
			p_2 + 
				h_{0} q_0\right)^{2}}}{2 \pi \left(g_{2}^{2} + 
				h_{0}^{2}\right)} - 
		\frac{h_{0} \sqrt{p_2^{2} + q_0^{2}}}{2 \pi \sqrt{g_{2}^{2} + 
		h_{0}^{2}}} + 
		\frac{h_{0}}{2 \pi \sqrt{g_{2}^{2} + h_{0}^{2}}} \nonumber\\&- 
		\frac{q_0 
			\operatorname{acos}{\left(\frac{g_{2} p_2}{\sqrt{g_{2}^{2} + 
			h_{0}^{2}} 
					\sqrt{p_2^{2} + q_0^{2}}} + \frac{h_{0} 
					q_0}{\sqrt{g_{2}^{2} + h_{0}^{2}} 
					\sqrt{p_2^{2} + q_0^{2}}} \right)}}{2 \pi} + 
					\frac{q_0}{4} + 
		\frac{\operatorname{acos}{\left(\frac{h_{0}}{\sqrt{g_{2}^{2} + 
		h_{0}^{2}}} 
				\right)}}{2 \pi} - \frac{1}{4},\\
		\text{(II)}\quad 0 &= - \frac{h_{0} 
		\operatorname{acos}{\left(\frac{g_{2} 
					p_2}{\sqrt{g_{2}^{2} + h_{0}^{2}} \sqrt{p_2^{2} + 
					q_0^{2}}} + \frac{h_{0} 
					q_0}{\sqrt{g_{2}^{2} + h_{0}^{2}} \sqrt{p_2^{2} + 
					q_0^{2}}} \right)}}{2 
			\pi} + \frac{h_{0}}{4} - \frac{q_0 \sqrt{g_{2}^{2} + 
			h_{0}^{2}}}{2 \pi 
			\sqrt{p_2^{2} + q_0^{2}}} - \frac{q_0}{2 \pi} + 
			\frac{q_0}{4} 
		\nonumber\\ &- 
		\frac{q_0 |p_2|}{2 \pi \left(p_2^{2} + q_0^{2}\right)} + 
		\frac{q_0 \sqrt{\left(g_{2} p_2 + h_{0} q_0\right)^{2}}}{2 \pi 
			\left(p_2^{2} + q_0^{2}\right)} + \frac{q_0}{2 \pi 
			\sqrt{p_2^{2} + 
				q_0^{2}}}+ 
				\frac{\operatorname{acos}{\left(\frac{q_0}{\sqrt{p_2^{2} + 
						q_0^{2}}} \right)}}{2 \pi} - \frac{1}{4}.
	\end{align*}
	In addition, by $[d^{-\frac{1}{2}}]\partder{4}$,
	\begin{align*}
		\text{(III)}\quad 0 &=- \frac{g_{2}}{2 \pi} + \frac{g_{2}}{4} - 
		\frac{g_{2} 
			|g_2|}{2 \pi 
			\left(g_{2}^{2} + h_{0}^{2}\right)} + \frac{g_{2} \sqrt{\left(g_{2} 
			p_2 + 
				h_{0} q_0\right)^{2}}}{2 \pi \left(g_{2}^{2} + 
				h_{0}^{2}\right)} - 
		\frac{g_{2} \sqrt{p_2^{2} + q_0^{2}}}{2 \pi \sqrt{g_{2}^{2} + 
		h_{0}^{2}}} + 
		\frac{g_{2}}{2 \pi \sqrt{g_{2}^{2} + h_{0}^{2}}} \\&- \frac{p_2 
			\operatorname{acos}{\left(\frac{g_{2} p_2}{\sqrt{g_{2}^{2} + 
			h_{0}^{2}} 
					\sqrt{p_2^{2} + q_0^{2}}} + \frac{h_{0} 
					q_0}{\sqrt{g_{2}^{2} + h_{0}^{2}} 
					\sqrt{p_2^{2} + q_0^{2}}} \right)}}{2 \pi} + 
					\frac{p_2}{4}.
	\end{align*}
	The system of the FPS equations is symmetric under $(g_2, 
	h_0)\leftrightarrow (p_2, q_0)$, and so we get the following symmetrized 
	version of (III),
	\begin{align*}
		\text{(IV)}\quad 0 &= - \frac{g_{2} 
		\operatorname{acos}{\left(\frac{g_{2} 
					p_2}{\sqrt{g_{2}^{2} + h_{0}^{2}} \sqrt{p_2^{2} + 
					q_0^{2}}} + \frac{h_{0} 
					q_0}{\sqrt{g_{2}^{2} + h_{0}^{2}} \sqrt{p_2^{2} + 
					q_0^{2}}} \right)}}{2 
			\pi} + \frac{g_{2}}{4} - \frac{p_2 \sqrt{g_{2}^{2} + 
			h_{0}^{2}}}{2 \pi 
			\sqrt{p_2^{2} + q_0^{2}}} - \frac{p_2}{2 \pi} \nonumber \\&+ 
		\frac{p_2}{4} - 
	\frac{p_2 |p_2|}{2 \pi \left(p_2^{2} + 
	q_0^{2}\right)}+ \frac{p_2 \sqrt{\left(g_{2} p_2 + h_{0} q_0\right)^{2}}}{2 
	\pi \left(p_2^{2} + q_0^{2}\right)} + \frac{p_2}{2 \pi 	\sqrt{p_2^{2} + 
	q_0^{2}}}.
	\end{align*}
\end{enumerate}

7. Our next goal is prove $p_2=g_2 = 0$. This step is somewhat more involved. 
Recall that for the family described in \pref{ex: fpsapp} 
$h_0 = 1/2 = -q_0$, and so the following expressions are well-defined,
\begin{align*}
	x &= - \frac{1}{2 \pi} + \frac{1}{4} - \frac{\abs{g_{2}}}{2 \pi 
		\left(g_{2}^{2} 
		+ h_{0}^{2}\right)} + \frac{\sqrt{\left(g_{2} p_2 + h_{0} 
			q_0\right)^{2}}}{2 \pi \left(g_{2}^{2} + h_{0}^{2}\right)} - 
	\frac{\sqrt{p_2^{2} + q_0^{2}}}{2 \pi \sqrt{g_{2}^{2} + h_{0}^{2}}} + 
	\frac{1}{2 \pi \sqrt{g_{2}^{2} + h_{0}^{2}}},\\
	y &=  - \frac{1}{2 \pi} + \frac{1}{4} - \frac{\abs{p_2}}{2 \pi 
		\left(p_2^{2} + q_0^{2}\right)} + \frac{\sqrt{\left(g_{2} p_2 + 
			h_{0} 
			q_0\right)^{2}}}{2 \pi \left(p_2^{2} + q_0^{2}\right)} - 
	\frac{\sqrt{g_{2}^{2} + h_{0}^{2}}}{2 \pi \sqrt{p_2^{2} + 
			q_0^{2}}} + \frac{1}{2 \pi 
		\sqrt{p_2^{2} + q_0^{2}}},\\
	z &= - \frac{\operatorname{acos}{\left(\frac{g_{2} p_2}{\sqrt{g_{2}^{2} 
					+ h_{0}^{2}} \sqrt{p_2^{2} + q_0^{2}}} + \frac{h_{0} 
				q_0}{\sqrt{g_{2}^{2} 
					+ h_{0}^{2}} \sqrt{p_2^{2} + q_0^{2}}} \right)}}{2 \pi} 
	+ \frac{1}{4}.
\end{align*}
Equations (I-IV) now read:
\begin{align*}
	\mathrm{(I)}\quad 0&= h_{0} x + q_0 z + 
	\frac{\operatorname{acos}{\left(\frac{h_{0}}{\sqrt{g_{2}^{2} + h_{0}^{2}}} 
			\right)}}{2 \pi} - \frac{1}{4},\\
	\mathrm{(II)}\quad 0&=h_{0} z + q_0 y + 
	\frac{\operatorname{acos}{\left(\frac{q_0}{\sqrt{p_2^{2} + q_0^{2}}} 
			\right)}}{2 \pi} - \frac{1}{4}
	,\\
	\mathrm{(III)}\quad 0&=g_{2} x + p_2 z
	,\\
	\mathrm{(IV)}\quad 0&=g_{2} z + p_2 y.
\end{align*}
Combining (I) and 
(III) (resp. (II) and (IV)) by solving (I) for $x$ (resp. solving (II) for $y$) 
and substituting yields,
\begin{align*}
	\mathrm{(A)} \quad 0 &= \frac{-g_{2}}{h_0}\left(q_0 z + 
	\frac{\operatorname{acos}{\left(\frac{h_{0}}{\sqrt{g_{2}^{2} + h_{0}^{2}}} 
			\right)}}{2 \pi} - \frac{1}{4}\right) + p_2 z,	\\
	\mathrm{(B)} \quad 0&= g_{2} z -  \frac{p_2}{q_0}\left(h_{0} z  + 
	\frac{\operatorname{acos}{\left(\frac{q_0}{\sqrt{p_2^{2} + q_0^{2}}} 
			\right)}}{2 \pi} - \frac{1}{4}\right).
\end{align*}
Summing (A) and (B) we have,
\begin{align*}
	0&= \left(p_2+g_2 -\frac{g_2q_0}{h_0} - \frac{p_2h_0}{q_0}\right)z
	-\frac{g_{2}}{h_0}\left(\frac{\operatorname{acos}{\left(\frac{h_{0}}{\sqrt{g_{2}^{2}
					+ h_{0}^{2}}} \right)}}{2 \pi} - \frac{1}{4}\right)
	-\frac{p_2}{q_0}\left(\frac{\operatorname{acos}{\left(\frac{q_0}{\sqrt{p_2^{2}
					+ q_0^{2}}} \right)}}{2 \pi} - \frac{1}{4}\right).
\end{align*}
For the derivation so far to be valid only $h_0,q_0\neq0$ is 
required. Plugging-in $h_0 = 1/2 = -q_0$, the above becomes
\begin{align} \label{eqn:h_and_j_halfs}
	0&= \frac{g_{2} \operatorname{acos}{\left(\frac{1}{2 \sqrt{g_{2}^{2} + 
					\frac{1}{4}}} \right)}}{\pi} - \frac{g_{2}}{2} - 
				\frac{p_2 
		\operatorname{acos}{\left(- \frac{1}{2 \sqrt{p_2^{2} + \frac{1}{4}}} 
			\right)}}{\pi} + \frac{p_2}{2}.
\end{align}
The function 
\begin{align*}
	f(x) = \frac{x \operatorname{acos}{\left(\frac{1}{2 \sqrt{x^{2} + 
					\frac{1}{4}}} \right)}}{\pi} - \frac{x}{2}
\end{align*}
is injective, and so $g_2=p_2$ by \pref{eqn:h_and_j_halfs}. Plugging 
in this into (A) yields $f(g_2) =0$. Since $f(0)=0$ we have, by 
the injectivity of $f$ again, that $g_2=0$, hence $h_2=0$ as well. 
Backward substitution then gives $c_4 = 2, ~f_4 = 1, e_6=0$. The FPS 
equations encountered in the reminder of the derivation are simpler.

\begin{enumerate}[leftmargin=*]
	\setcounter{enumi}{7}

	\item We have $[d^{-\frac{1}{4}}] \partder{5} = - 
	\frac{h_{1}}{2 \pi} + 
	\frac{h_{1}}{4} 
	- 
	\frac{q_1}{4} + \frac{q_1}{2 \pi}$, hence $h_1=q_1$. Therefore, 
	$[d^{-\frac{1}{4}}]\partder{1} = \frac{c_{5}}{2 
	\pi} - \frac{h_{1}}{2 \pi} + 
	\frac{q_1}{2 
		\pi}$ implies $c_5 = 0$.
	
	\item By $[d^{-\frac{1}{2}}] \partder{3} = 
	\frac{f_{6}}{4} + \frac{h_{2}}{4} + 
	\frac{q_2}{4}$ 
	and $[d^{-\frac{1}{2}}] \partder{5}= - 
	\frac{c_{6}}{2 \pi} + 
	\frac{f_{6}}{4} + \frac{h_{2}}{2}$, hence
	$f_6 = -h_2 - q_2$ and $c_{6} = \frac{\pi h_{2}}{2} - \frac{\pi q_2}{2}$.
	
	\item By $[d^{-\frac{3}{4}}] \partder{2} = 
	\frac{e_{7}}{4} + \frac{g_{3}}{4} + 
	\frac{p_3}{4}, ~ 
	[d^{-\frac{3}{4}}] \partder{3} = \frac{f_{7}}{4} 
	+ \frac{g_{3}}{2 \pi} + 
	\frac{h_{3}}{4} 
	- 
	\frac{p_3}{2 \pi} + \frac{q_3}{4}, 
	[d^{-\frac{3}{4}}] \partder{4}  =\frac{e_{7}}{4} 
	+ 
	\frac{g_{3}}{2}$ and $[d^{-\frac{3}{4}}] 
	\partder{6}  =\frac{e_{7}}{4} + 
	\frac{p_3}{2}$, $e_7 = -2p_3,~f_7 = -h_3 - j_3$ and $g_3 = p_3$.
	
	\item By $[d^{-\frac{3}{4}}] \partder{5}$, $c_{7} 
	= \frac{\pi h_{3}}{2} - 4 
	p_3 - \frac{\pi q_3}{2}$, and by $[d^{-1}] \partder{2}$, $e_{8} = - g_{4} - 
	p_4 + \frac{4}{\pi}$.
	
	\item We now have
	\begin{align*}
		[d^{-1}] \partder{4} &= \frac{g_{4}}{4} + \frac{2 p_3 
		\abs{p_3}}{\pi} 
		- 
		\frac{p_4}{4} - \frac{1}{2} + \frac{3}{2 \pi},\\ 
		[d^{-1}] \partder{6} &= - \frac{g_{4}}{4} + \frac{2 p_3 
		\abs{p_3}}{\pi} + 
		\frac{p_4}{4} - \frac{1}{2} + \frac{1}{2 \pi}.
	\end{align*}
	Summing the two equations gives
	\begin{align*}
		\frac{4 p_3 \abs{p_3}}{\pi} - 1 + \frac{2}{\pi} = 0.
	\end{align*}
	The equation has a single root at $p_3 =g_3 = \frac{\sqrt{-2 + \pi}}{2}$. 
	Backward substitution then gives $e_7 = - \sqrt{-2 + \pi}$.
	
	\item The expression $[d^{-\frac{1}{2}}] 
	\partder{1}$ depends on $g_3$ which has 
	just been determined. Re-evaluating, we have $h_{2} = q_2 + 1$, and so $c_6 
	= 	\frac{\pi}{2}$. 
	
	\item By $[d^{-1}] \partder{4}$, $g_{4} = p_4 - \frac{2}{\pi}$, by 
	$[d^{-1}]\partder{3},~f_{8} = - h_{4} - q_4 + \frac{4}{\pi^{2}}$, and by 
	$[d^{-\frac{3}{4}}]\partder{1}$, 
	
	$$h_{3} = \frac{- 4 \pi p_4 \sqrt{-2 + \pi} - 
	q_3(\pi^{2} + 2 \pi) + 4 \sqrt{-2 + \pi}}{\pi \left(2 - \pi\right)}.$$ 
	\item Now, 
	\begin{align*}
		[d^{-1}]\partder{5} &= 	- \frac{c_{8}}{2 \pi} + \frac{h_{4}}{4} - 
		\frac{2 
		p_4}{\pi} - 
		\frac{4 q_1 \sqrt{-2 + \pi}}{3 \pi} + \frac{2 q_1 \sqrt{-2 + 
		\pi}}{3} - 
		\frac{q_4}{4} - \frac{1}{2} - \frac{1}{2 \pi} + \frac{\pi}{8} + 
		\frac{7}{\pi^{2}}, \\
		[d^{-1}]\partder{7} &= 	\frac{c_{8}}{2 \pi} - \frac{h_{4}}{4} + \frac{2 
		p_4}{\pi} - 
		\frac{4 q_1 \sqrt{-2 + \pi}}{3 \pi} + \frac{2 q_1 \sqrt{-2 + 
		\pi}}{3} + 
		\frac{q_4}{4} - \frac{5}{\pi^{2}} - \frac{\pi}{8} + \frac{1}{2 \pi}.
	\end{align*}
	Summing the two equations above, we get
	\begin{align*}
		4 q_1 \sqrt{-2 + \pi} \left(\frac{1}{3} - \frac{2}{3 \pi}\right) - 
		\frac{1}{2} + \frac{2}{\pi^{2}} = 0,
	\end{align*}
	hence $q_1=h_1 = \frac{6 + 3 \pi}{8 \pi \sqrt{-2 + \pi}}$ 
	(recall that $q_1=h_1$). Consequently, $f_5 = - \frac{6 + 3 \pi}{4 \pi 
	\sqrt{-2 + \pi}}$.
	
	\item The procedure may be further iterated by observing, e.g., that 
	$[d^{-1}]\partder{5}$ implies $c_8 = \frac{\pi h_{4}}{2} - 4 p_4 - 
	\frac{\pi q_4}{2} - \frac{\pi}{2} - 1 + \frac{\pi^{2}}{4} + 
	\frac{12}{\pi}$ and so on, if additional coefficients are needed. 
	
\end{enumerate}

For type II critical points, the equations corresponding to Equations (I-IV) 
above are different, and we have not been able to solve them exactly. 
Rather, Newton-Raphson method was used to obtain numerical 
estimates. The same 
procedure was then applied iteratively, with the aid of numerical methods, 
giving estimates for higher-order terms. The estimates obtained 
through this process match with those obtained by the method 
described in \pref{sec:fps_der}. Other choices of types, $k$, $d$ and isotropy 
were addressed similarly.

\subsection{Expected Initial Value} \label{sec:expiv}
Bounding $\EE_W[\cL(W)]$ follows by a straightforward computation of the 
expected loss. We first derive explicit expressions for the terms used in 
computation.

\begin{itemize}[label={}, leftmargin=*]
\item 
\begin{flalign*}
	\EE_{{(\x,\w)\sim\cN(0,I_d)^{\otimes 
				2}}}\brk{\sigma^2\prn{{{\inner{\w,\x}}}}} 
	&= \EE_{\w\sim\cN(0,I_d)} \frac{\nrm{\w}^2}{2} =\frac{d}{2},&
\end{flalign*}	

\item 
\begin{flalign*}
	\EE_{{\x\sim\cN(0,I_d)}}\brk{\sigma\prn{{\inner{\v_2, 
	\x}}} 
		\sigma\prn{{\inner{\v_1, \x}}}} &= 
	\begin{cases}
		1/2 & \v_1=\v_2,\\
		1/(2\pi) & \v_1\neq \v_2,
	\end{cases}&
\end{flalign*}
\item 
\begin{flalign*}
	\EE_{\w\sim \cN(0, I_d)}\brk{\sigma\prn{{\inner{\w, 
	\x}}}}&= 
\frac{1}{2}\EE_{\w\sim \cN(0, 
I_d)}\brk{\sigma\prn{{\inner{\w,
				\x}}}~|~\inner{\w, \x}\ge0}
+\frac{1}{2}\EE_{\w\sim \cN(0, 
I_d)}\brk{\sigma\prn{{\inner{\w,
				\x}}}~|~\inner{\w, \x}<0}&\\
&= 	\frac{1}{2}\EE_{\w\sim \cN(0,I_d)}\brk{{{\inner{\w, 
				\x}}}~|~\inner{\w, \x}\ge0}&\\
&= 	\frac{1}{2}\EE_{\w\sim 
\cN(0,I_d)}\brk{{{{\w}}}~|~\inner{\w, 
	\x}\ge0}^\top 
\x&\\
&= 	\frac{\nrm{\x}}{\sqrt{2\pi}},
\end{flalign*}	

\item 
\begin{flalign*}
	\EE_{\w\sim \cN(0, I_d)}\brk{\sigma\prn{{\inner{\w, 
	\x}}}}&= 
	\frac{1}{2}\EE_{\w\sim \cN(0, 
	I_d)}\brk{\sigma\prn{{\inner{\w,
					\x}}}~|~\w^\top \x\ge0}
	+\frac{1}{2}\EE_{\w\sim \cN(0, 
	I_d)}\brk{\sigma\prn{{\inner{\w, 
					\x}}}~|~\inner{\w, \x}<0}&\\
	&= 	\frac{1}{2}\EE_{\w\sim \cN(0,I_d)}\brk{{{\inner{\w,
					\x}}}~|~\inner{\w, \x}\ge0}&\\
	&= 	\frac{1}{2}\EE_{\w\sim 
	\cN(0,I_d)}\brk{{{{\w}}}~|~\inner{\w,
		\x}\ge0}^\top \x&\\
	&= 	\frac{\nrm{\x}}{\sqrt{2\pi}},
\end{flalign*}
\item 
\begin{flalign*}		
	\EE_{\nrm{\x}=1}\brk{\sigma\prn{{\inner{\v, \x}}}}&= 
	\frac{\nrm{\v}}{2}
	\EE_{\theta}\brk{{{\inner{\v/\nrm{\v}, 
	\x}}}~|~\inner{\v/\nrm{\v},	\x}\ge0}&\\
	&= \frac{\nrm{\v}}{2}\EE_{\nrm{\x}=1}\brk*{x_1~|~x_1\ge0}&\\
	&= 	\frac{\nrm{\v}}{2}\frac{2}{d\text{Beta}((n+1)/2,1/2)}&\\
	&= 	\frac{\nrm{\v}}{d\text{Beta}((n+1)/2,1/2)},
\end{flalign*}		

\item 

\begin{flalign*}		
	\EE_{_{\x\sim\cN(0,I_d)}^{\w\sim \cN(0, I_d)}}&
	\brk{\sigma\prn{{\inner{\w, \x}}}\sigma\prn{{\inner{\v,
					\x}}}} \\&=
	\EE_{\x\sim\cN(0,I_d)}\brk*{ \frac{1}{\sqrt{2\pi}}\nrm{\x}  
		\sigma\prn{{\inner{\v, \x}}}}&\\
	&= \EE_{r} 
	\EE_{\nrm{\btheta}=1}\brk*{\frac{1}{\sqrt{2\pi}}r\nrm{\btheta}
		\sigma\prn{{\inner{\v, r\btheta}}}}&\\
	&= \EE_{r} \frac{v}{\sqrt{2\pi}}r^2 \EE_{\nrm{\btheta}=1}\brk*{	
		\sigma\prn{{\inner{\v, \btheta}}}}&\\
	&= 
	\frac{1}{\sqrt{2\pi}}\frac{\nrm{\v}}{d\text{Beta}((d+1)/2,1/2)}\EE_{r}
	r^2 &\\
	&= \frac{\nrm{\v}}{\sqrt{2\pi}\text{Beta}((d+1)/2,1/2)},&
\end{flalign*}
\item 
\begin{flalign*}	
	\EE_{{(\x,\v,\w)\sim\cN(0,I_d)^{\otimes	
	3}}}&\brk{\sigma\prn{{\inner{\w,
					\x}}}\sigma\prn{{\inner{\v, \x}}}}\\ 
	&= \frac{\EE_{\v}\nrm{\v}}{\sqrt{2\pi}\text{Beta}((d+1)/2,1/2)}	&\\
	%
	&= 
	\frac{1}{\sqrt{\pi}\text{Beta}((d+1)/2,1/2)}\frac{\Gamma((d+1)/2)}{\Gamma(d/2)}
	&\\
	&=\frac{1}{\pi\text{Beta}(d/2,1)}&\\
	&=\frac{d}{2\pi}.
\end{flalign*}
\end{itemize}
Therefore,
\begin{align*}
	\EE_{_{W\sim\cN\prn*{0_{d\times 
	d},I_{d^2}}}^{\x\sim\cN(0,I_d)}}&\brk{ 
		(\bones^\top 
		\sigma\prn*{\frac{1}{\sqrt{d}}W\x}- \bones^\top 
		\sigma(V\x))^2   } \\
	&=\EE\brk{ 
		d \frac{1}{d}\sigma^2(\inner{\w,\x}) + \frac{1}{d} 
		d(d-1)\sigma(\inner{\w_1,\x})\sigma(\inner{\w_2,\x}) 
		-\frac{1}{\sqrt{d}}2 d^2 
		\sigma(\inner{\w,\x})\sigma(\inner{\v,\x}) +  d 
		\sigma^2(\inner{\v, \x}) \\
		&+ 
		d(d-1)\sigma(\inner{\v_1,\x})\sigma(\inner{\v_2,\x}) 
		 }\\
	&= \frac{d}{2} +(d-1)\frac{d}{2\pi} 
	-  \frac{2 d^{3/2}}{\sqrt{2\pi } \text{Beta}((d+1)/2,1/2)} + 
	\frac{d}{2} + 
	d(d-1)\frac{1}{2\pi}\\
	&= d +\frac{d(d-1)}{\pi} 
	-  \frac{\sqrt{2} d^{3/2} \Gamma(d/2+1)}{\sqrt{\pi } 
		\Gamma((d+1)/2)\Gamma(1/2)}\\
	&\le d +\frac{d(d-1)}{\pi} 
	-  \frac{\sqrt{2} d^{3/2}}{\pi} \prn*{\frac{d}{2}}^{1/2}\\
	&= \prn*{1-\frac{1}{\pi}}d,
\end{align*}
where the penultimate transition uses Gautschi's inequality. 
The same inequality also gives the following lower bound,
\begin{align*}
	\EE\brk{(\bones^\top \sigma\prn*{\frac{1}{\sqrt{d}}W\x}- 
	\bones^\top 
		\sigma(V\x))^2   } 	&\ge \prn*{1-\frac{1}{\pi}}d +	
		\frac{1}{\pi} (d^2 
	- 
	\sqrt{2}d^{3/2}(d/2+1)^{1/2})\\
	&= \prn*{1-\frac{1}{\pi}}d +	
	\frac{d^{2}\prn*{1-\sqrt{1+\frac{2}{d}}}}{\pi} 
	\\
	&\ge \prn*{1-\frac{1}{\pi}}d +	\frac{d^{2}\prn*{1- (1+1/d) }}{\pi} 
	\\
	&= \prn*{1-\frac{2}{\pi}}d,
\end{align*}
with the penultimate transition following by the first-order Taylor 
expansion of $\sqrt{1+x}$.

\subsection{Adding more than two neurons: beyond non-degenerate critical  
points}\label{sec:fossil}

When we add neurons to a shallow network, new critical points appear and old 
critical points become simplices with singular Hessian spectrum (at points 
where the Hessian is defined). This phenomenon is well-known and not restricted 
to ReLU networks. We shall 
refer to this process here as ``fossilization'', as the set 
of (connected) fossils together, with the discrete critical points, encodes 
information about the number of additional neurons involved,
and how they were added.
Thus the fossil record generated when $p>1$ neurons are added simultaneously may be less informative than that generated when $p$ neurons are added one-by-one. Moreover, as we show, symmetry 
plays a significant role in the description of the fossilized sets;  even if the target $V$ is asymmetric.

The critical points giving the global minimum fossilize when neurons are added. Recall 
that $\ploss$ is always $S_k$-invariant, where $S_k$ is the group of row permutations of $M(k,d)$.  If $d = k$, we add the superscript $r$ (resp.~$c$) to emphasize row (resp.~column) permutations.
In our setting, if $k=d$, $V = I_d$, the $d!$ points in the 
$S^r_d$-orbit of $V$ will be non-degenerate critical points of $\ploss$ 
giving the global minimum zero: these will be the only points in $M(d,d)$ that 
give the global minimum. If we add $p \ge 1$ neurons, 
the discrete $S_d^r$-orbit of $V$ is replaced by a $p$-dimensional connected $S_{d+p}$-invariant
simplicial complex $Z\subset M(p+d,d)$ consisting of all points giving the global minimum. Necessarily, the Hessian, where defined on $Z$, will have zero eigenvalues; that does not 
preclude $Z$ from being an attractor under gradient descent. Suppose instead that $\mathfrak{c} \in M(k,d)$, $k \ge d$, is any (non-degenerate) critical point of $\ploss$. 
The addition of a $p$ neurons will replace $\mathfrak{c}$ by a connected 
$p$-dimensional simplex, invariant by the action of $S_d^r$. Often (not always) 
many new non-degenerate critical points will be created as biproducts of the 
fossilization process.

For completeness, we give precise statements and proof of these results, 
starting with the case when $\mathfrak{c}=V$ and $k = d$. We start with an 
extension of the result on the uniqueness of critical points defining global 
minima \cite[Prop.~4.14]{ArjevaniField2020} to the over-parameterized case. See 
\cite{safran2021effects} for related results and discussions.

Assume $k \ge d$, set $m = k-d$, and $\Gamma = S_k \times S_d$. Let $S_m = \{e_d\} \times S_m$ denote the subgroup of $S_k$ permuting the last $m$ rows of matrices in $M(k,d)$.
Define $\Delta_m S_d   =  \{(hg,g) \dd g S_d,\, h  \in S_m\}$. Note that
$\Delta_m S_d \approx \Delta S_d \times S_m$.

Let $\mathfrak{K}^\star$ denote the set of all partitions of $[k]$ such that
each $\mathcal{K} \in \mathfrak{K}^\star$ has exactly $d$ parts, $K_1,\cdots K_d$ and $j \in K_j$, for all $j \in \is{d}$.
If $j \in \is{d}$, then
\[
K_j \cap \is{d} = \{j\}, \; \;\text{and } K_j \smallsetminus \{j\} \subset \is{k} \smallsetminus \is{d}.
\]
Clearly, $ 1\le |K_j| \le m+1 $ for all $j \in \is{d}$.

If $\mathcal{K} \in \mathfrak{K}^\star$, let $M_{\mathcal{K}}=[k_{ij}]\in M(k,d)$ be the
matrix defined by
\begin{eqnarray}\label{eq: d0}
k_{ij} & = & 0, \; i \notin K_j \\
\label{eq: d1}
 & = & 1,\;i \in K_j
\end{eqnarray}
For $j \in \is{d}$, define $\Delta_j(\mathcal{K}) \subset \real^k$ by
\[
\Delta_j(\mathcal{K}) = \{(t_1,\cdots,t_k) \in \prod_{i \in K_j} [0,k_{ij}] \dd \sum_{i \in K_j} t_{i} = 1\}
\]
and, viewing $M(k,d)$ as $(\real^k)^d$, define
\[
\Delta(\mathcal{K}) = \prod_{j \in [d]} \Delta_j(\mathcal{K})\subset M(k,d).
\]
Clearly $\Delta(\mathcal{K})$ is a simplicial complex of dimension $m$ and if $m = 0$, $\Delta(\mathcal{K}) = \{V\}$. If $\boldsymbol{\delta} \in \Delta(\mathcal{K})$, then
\[
\boldsymbol{\delta}^\Sigma = V^\Sigma = \mathcal{I}_{1,d},
\]
where for $W\in M(k,d)$, $W^\Sigma$ is the $1 \times d$-row matrix defined by the column sums of $W$.
Define
\begin{equation}
\boldsymbol{\Delta}(k,d) = \bigcup_{\mathcal{K} \in \mathfrak{K}^\star} \Delta(\mathcal{K}).
\end{equation}

Suppose $W \in M(k,d)$ and $\boldsymbol{\delta}\in \Delta(\mathcal{K})$. Define $\WW_{\boldsymbol{\delta}} \in M(k,d)$ by
\begin{eqnarray*}
\ww^{\boldsymbol{\delta}}_i&=& \delta_{ii}\ww_i, \; i\in 
\is{d}\\
&=&\ww_i + \sum_{j \in \is{d}}\delta_{ij}\ww_j, \; i> d
\end{eqnarray*}
Observe that
\[
W^\Sigma = W_{\boldsymbol{\delta}}^\Sigma.
\]
The case of most interest will be when $W = V$ and so the last $m$ rows of $W$ will be zero.

\begin{lemma}\label{lem:sc}
(Notation and assumptions as above.)
\begin{enumerate}
\item For all $\mathcal{K} \in \mathfrak{K}^\star$, $V \in \Delta(\mathcal{K})$.
\item
If $\mathcal{K}, \mathcal{J} \in \mathfrak{K}^\star$, $\mathcal{K} \ne \mathcal{J}$, then $\Delta(\mathcal{K})\cap \Delta(\mathcal{J})$ is a simplicial complex
which is the union of the common vertices and faces of  $\Delta(\mathcal{K})$ and $\Delta(\mathcal{J})$.
\item $\boldsymbol{\Delta}(k,d)$ is a connected simplicial complex of dimension $m$.
\item $\Delta_m S_d(\boldsymbol{\Delta}(k,d)) = \boldsymbol{\Delta}(k,d)$ and $\Delta_m S_d$ is the maximal subgroup of $\Gamma$ with this property.
\item If $g \in S_d^r \cup S_d^c$, $g\ne e$, then $g \boldsymbol{\Delta}(k,d) \cap \Gamma V = g V \ne V$.
\item $\mathcal{L}(W) = 0$ if $W \in \boldsymbol{\Delta}(k,d)$.
\end{enumerate}
\end{lemma}
\proof Statements (1--3) are all immediate from the 
definitions and the proofs of (4,5) are
straightforward and omitted.
It remains to prove (6). Recall that 
\[
\mathcal{L}(W) = \frac{1}{2}\sum_{i,j \in [k]} 
f(\wwi{i},\wwi{j}) - \sum_{i\in \is{k},j \in \is{d}} 
f(\wwi{i},\vvi{j}) +  \frac{1}{2}\sum_{i,j \in 
\is{d}}f(\vvi{i},\vvi{j}),
\]
where
\begin{enumerate}
\item If $\v,\w\in \real^d$ are non-zero and we set 
$\theta_{\w,\v} = \cos^{-1}\left(\frac{\langle 
\w,\v\rangle}{\|\w\|\|\v\|}\right)$, then
\[ 
f(\w,\v)  = \frac{1}{2\pi} 
\|\w\|\|\v\|\big(\sin(\theta_{\w,\v}) + 
(\pi-\theta_{\w,\v})\cos(\theta_{\w,\v})\big)
\]
\item $f(\w,\v) = 0$ iff either $\v$ or $\w$ is zero or 
$\theta_{\w,\v} = \pi$.
\end{enumerate}
Clearly, $f$ is positively homogeneous:
\[
f(a\w,b\v) = ab f(\w,\v), \; a,b \ge 0, \;\w,\v \in \real^d.
\]
If $W \in \boldsymbol{\Delta}(k,d)$, then there exist  $\mathcal{K}\in \mathfrak{K}^\star$ and $\boldsymbol{\delta} \in \Delta(\mathcal{K})$ such that
$W = V_{\boldsymbol{\delta}}$---$\vvi{j}$ is zero for $j > 
d$. The result follows from the positive homogeneity of $f$ 
and the formula for $\mathcal{L}$ in terms of $f$.  \qed

\begin{rem}
If $k = d$, $V$ is the natural choice for a critical point on the group orbit of critical points giving the global minima. When $k > d$, the natural choice---at least from a symmetry perspective---is
the set $\boldsymbol{\Delta}(k,d)$ which is invariant by $\Delta_m S_d $ (the isotropy group of $V$). It follows by (5) of the lemma that for all $g\in S_d \times S_d$, $g \boldsymbol{\Delta}(k,d)$
contains exactly one point in $\Gamma V$. 
\end{rem}

Define $ \boldsymbol{\Delta^\star} = \Gamma \boldsymbol{\Delta}(k,d)$ and note that $\Gamma \is{V} \subset \boldsymbol{\Delta^\star}$.

\begin{theorem}\label{thm: gm}
(Notation and assumptions as above.)
\begin{enumerate}
\item $\boldsymbol{\Delta^\star}$ is a $\Gamma$-invariant $m$-dimensional simplicial complex of $M(k,d)$.
\item $\boldsymbol{\Delta^\star}$ is connected.
\item $\mathcal{L}(W) = 0$ iff $W \in \boldsymbol{\Delta^\star}$.
\end{enumerate}
\end{theorem}
\proof By~\pref{lem:sc}, $\boldsymbol{\Delta}(k,d)$ is a connected $m$-dimensional simplicial complex and it follows easily that $\boldsymbol{\Delta^\star}$ is a $\Gamma$-invariant $m$-dimensional simplicial complex, proving (1).
(2) If $g \in \Gamma$, $g \boldsymbol{\Delta}(k,d) \cap \boldsymbol{\Delta}(k,d)$ may be empty. However, given any $g \in\Gamma$,
it is easy to choose a sequence  $g_1,\cdots g_{n+1} \in  S_k$ such that such that $g_1 = e$,
and $g_j \boldsymbol{\Delta}(k,d)  \cap g_{j+1}\boldsymbol{\Delta}(k,d)  \ne \emptyset$, $j \in \is{n}$, and
$g_{n+1}\boldsymbol{\Delta}(k,d) \cap g \boldsymbol{\Delta}(k,d)  \ne \emptyset$.  Hence $\Gamma \boldsymbol{\Delta}(k,d)  = \boldsymbol{\Delta^\star}$ is connected (see
\pref{ex: nc_ex} below for more detail). It is straightforward to check that
for all $g \in \Gamma$, and $\mathcal{K},\mathcal{J} \in \mathfrak{K}^\star$, $g \Delta(\mathcal{K}) \cap \Delta(\mathcal{J})$ is a simplicial complex (possibly empty) and from this it follows
that $\boldsymbol{\Delta^\star}$ is  a simplicial complex. \\
(3)  The `if' implication is immediate from the $\Gamma$-invariance of 
$\mathcal{L}$, the connectedness of  $\boldsymbol{\Delta^\star}$ and 
\pref{lem:sc}(6). For the converse, we show that if $\mathcal{L}(W) = 0$, then 
(a) $w_{ij} \in [0,1]$, $(i,j) \in \is{k}\times \is{d}$ and $\sum_{i\in\is{k}} 
w_{ij} = 1$, for all $j \in \is{d}$ (and so $W^\Sigma = 
\mathcal{I}_{1,d}$).
The remainder of the proof follows along the same lines as that of \cite[Prop.~4.14]{ArjevaniField2020} except that now for each $j \in \is{d}$ we have to allow for several rows
of $W$ being strictly positive multiples of $v^j$ since $k > d$.
\qed

\begin{example}\label{ex: nc_ex}
Suppose $m = 1$ and $d = 2$.   We claim that $\boldsymbol{\Delta^\star}$ is connected. Here the set $\mathfrak{K}^\star$ contains only two partitions: $K = \{\{1,3\},\{2\}\}$,
$J = \{\{1\},\{2,3\}\}$.  Hence there are two families of matrices in $\boldsymbol{\Delta}(3,2)$
\[
X(\alpha,\beta)=\left[\begin{matrix}
\alpha & 0 \\
0 & 1 \\
\beta & 0
\end{matrix}\right], \quad 
Y(\gamma,\delta)= \left[\begin{matrix}
1 & 0 \\
0 & \gamma \\
0 & \delta
\end{matrix}\right],
\]
where $\alpha + \beta = \gamma + \delta = 1$, $\alpha,\beta,\gamma,\delta \ge 0$.

Let $\boldsymbol{\Delta^{\star}_{0}}$ denote the connected component of 
$\boldsymbol{\Delta^\star}$ containing $X$ (that is, the arc $X(\alpha,\beta)$, 
$\alpha+\beta=1$, $\alpha,\beta \ge 0$).
Use the symbol $\sim$ is signify that two families intersect. For example, $X\sim Y$ since $X(1,0) = Y(1,0)$.
We claim
$(12)^rX \in \boldsymbol{\Delta^{\star}_{0}}$.
This follows since $X_1 = (13)^rX \sim X$, $X_2 = (12)^r X_1 \sim X_1$, $X_3 = (23)^r X_2 \sim X_2$ and $X_3 =(12)^rX$. It is easy to see that
$\boldsymbol{\Delta^\star}$ is isomorphic to a hexagon: 6 vertices, 6 edges and that $\Delta S_2(\boldsymbol{\Delta}(3,2))=\boldsymbol{\Delta}(3,2)$,
where $\Delta S_2 = \Gamma_V$.
See Figure~\ref{fig: hex} where the vertices, connecting edges and symmetries of $\boldsymbol{\Delta^\star}$ are shown.
\begin{figure}[h]
\centering
\includegraphics[width=\textwidth]{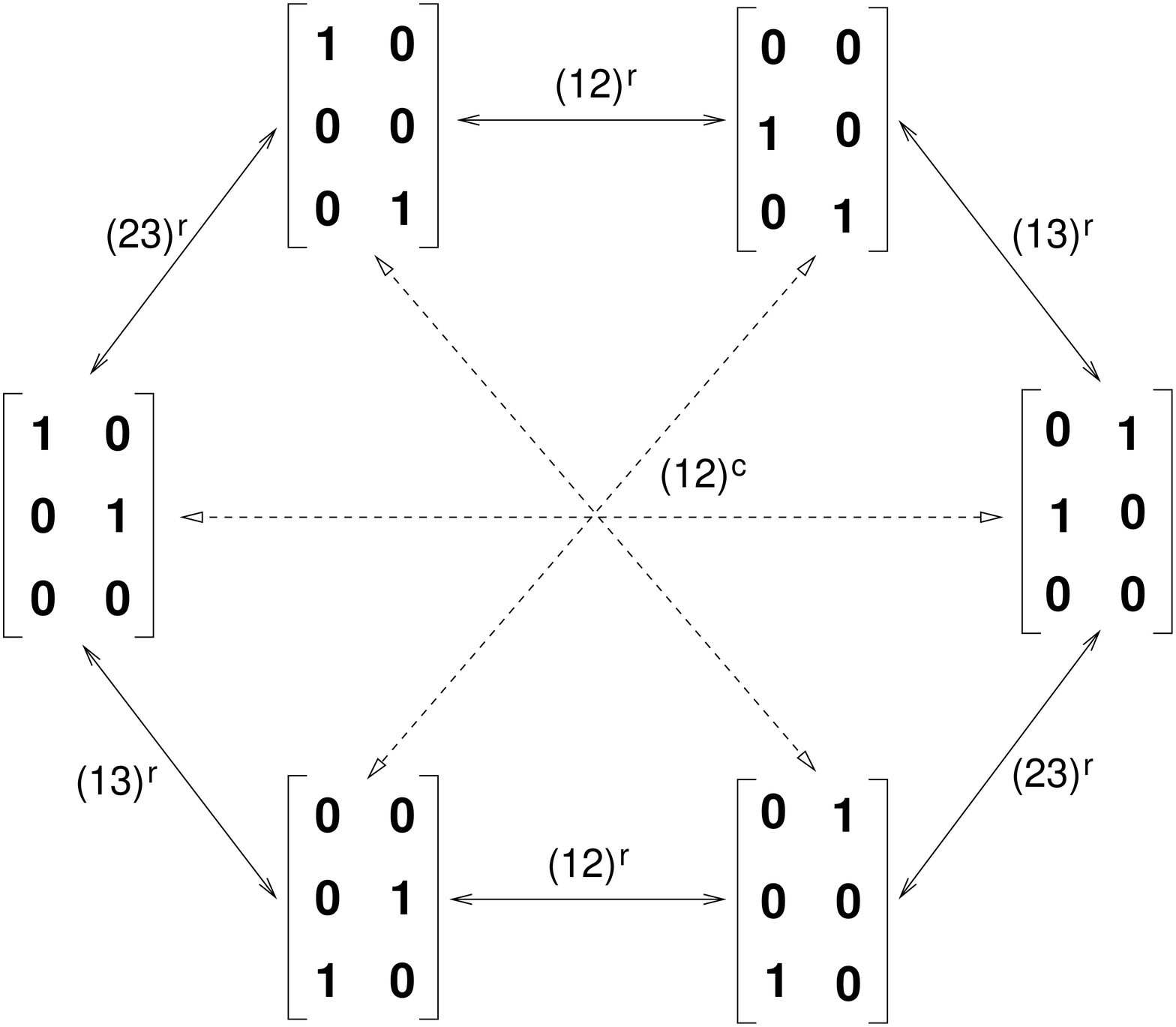}
\caption{The simplex $\boldsymbol{\Delta^\star}$  in case $k = 3, d= 2$. Connecting edges are shown using unbroken lines and are labelled by the row transposition that interchanges vertices.
The simplicies $\boldsymbol{\Delta}(3,2), (12)^c \boldsymbol{\Delta}(3,2)$ are both invariant by $\Delta S_2$. }
\label{fig: hex}
\end{figure}

The argument is general and applies when $k > d$---that is, when $V\in M(k,d)$ has at least one row of zeros. The connection can always be made through \emph{row} permutations.
\end{example}

\noindent {\bf Fossilization of critical points, general case.}
 
The phenomenon described above occurs when the network is 
over-parameterized.  In what follows we assume $V \in 
M(d,d)$ is
a matrix with no zero or parallel rows, and extend in the usual way to $V \in M(k,d)$, $k > d$. 

Suppose $W\in M(k,d)$ is a critical point of $\mathcal{L}$: 
in particular, assume that $\mathcal{L}$ is $C^2$ 
at $W$ and so
$W$ has no zero rows. Typically, we assume that  $W$ is non-degenerate: all the eigenvalues of the Hessian are non-zero.

Let $\bar{k} > k$ and
$\mathfrak{K}^\star$ denote the set of all partitions of $\is{\bar{k}}$ such that
each $\mathcal{K} \in \mathfrak{K}^\star$ has exactly $k$ parts, $K_1,\cdots K_k$ and $j \in K_j$, for all $j \in \is{k}$.

Just as we did previously, if  $\mathcal{K} \in \mathfrak{K}^\star$ and $j \in \is{k}$, we define the simplex $\Delta_j(\mathcal{K}) \subset \real^{\bar{k}}$, and
simplicial set $\Delta(\mathcal{K}) = \prod_{j\in \is{k}} \Delta_j(\mathcal{K})\subset M(\bar{k},k)$ of dimension $\bar{k}-k$.  Set
$\boldsymbol{\Delta}(\bar{k},k) = \cup_{\mathcal{K} \in  \mathfrak{K}^\star} \Delta(\mathcal{K})$.

Given $W\in M(k,d)$, $\mathcal{K} \in \mathfrak{K}^\star$ and $\boldsymbol{\delta} \in \Delta(\mathcal{K})$, define
$W_\delta \in M(\bar{k},d)$ by
\begin{eqnarray*}
\w^{\boldsymbol{\delta}}_i&=& \delta_{ii}\w_i, \; i\in 
\is{k}\\
&=&\sum_{j \in \is{k}}\delta_{ij}\w_j, \; i> k
\end{eqnarray*}
As we did above when $W = V$, we have $W_\delta^\Sigma = W^\Sigma$.
It follows from the definitions that if  
$\boldsymbol{\delta}\in \Delta (\mathcal{K})$, then 
$\sum_{\ell \in K_i} \ww_\ell^{\boldsymbol{\delta}}  = 
\w_i$, for all $i \in \is{k}$.

Given $W \in  M(k,d)$, define $\Delta(\bar{k},k)(\WW) = \{W_{\boldsymbol{\delta}} \dd \boldsymbol{\delta} \in \Delta(\bar{k},k)\}$ and
\[
\boldsymbol{\Delta^\star}(W)=\Gamma\Delta(\bar{k},k)(W)\subset M(\bar{k},d).
\]
Set $\boldsymbol{\Delta^{\star\star}}(W) = \boldsymbol{\Delta^\star}(W) \smallsetminus \partial \boldsymbol{\Delta^\star}(W)$

\begin{proposition}\label{proposition: gen}
(Notation and assumptions as above.)
\begin{enumerate}
\item $\boldsymbol{\Delta^\star}(W)$ is a connected $S_d^r$-invariant simplical complex of dimension $\bar{k}-k$.
\item $\mathcal{L}$ is $C^2$ at all points of ${\boldsymbol{\Delta}}^{\star\star}(W)$ and $\nabla \mathcal{L}| {\boldsymbol{\Delta}}^{\star\star}(W) \equiv 0$
\item $\mathcal{L}$ is constant on $\boldsymbol{\Delta^\star}(W)$.
\end{enumerate}
\end{proposition}
\proof The proof of (1) is similar to that of \pref{thm: gm} (1,2). Since $\boldsymbol{\Delta^\star}(W)$ is connected, and $\mathcal{L}$ is continuous,
(3) follows from (2) ((3) may be proved directly as in \pref{thm: gm} and then (2) follows using the
regularity of $\mathcal{L}$ on ${\boldsymbol{\Delta}}^{\star\star}(W)$). For completeness, we give a direct proof that $\grad{\mathcal{L}}$ vanishes on ${\boldsymbol{\Delta}}^{\star\star}(\WW)$.
By an obvious induction, we can reduce to showing that if the partition $\mathcal{K} \in \mathfrak{K}^\star$ satisfies $|K_i| = 1$, $i < k$ and $|K_k|-1 = n = \bar{k} -k \ge 1$,
then $\nabla \mathcal{L}| \Delta (\mathcal{K}) \smallsetminus \partial \Delta (\mathcal{K}) \equiv 0$. Set $m = n+1 = |K_k|$.

By standard results on $\nabla \ploss$~\cite{brutzkus2017globally}, $W$ is a critical point of $\mathcal{L}$ if for $i \in \is{k}$
\begin{equation}\label{eq: crystal}
\sum_{j\in \is{k}} 
\left(\frac{\|\wwi{j}\|\sin(\theta_{\wwi{i},\wwi{j}})}{\|\wwi{i}\|}\wwi{i}
 -\theta_{\wwi{i},\wwi{j}}\wwi{j}\right)-  
 \sum_{j\in 
 \is{d}}\left(\frac{\sin(\theta_{\wwi{i},\vvi{j}})}{\|\wwi{i}\|}\wwi{i}
  - \theta_{\wwi{i},\vvi{j}}\vvi{j}\right) + \pi\big(W - 
 V\big)^\Sigma=0
\end{equation}
Suppose that $U \in  \Delta (\mathcal{K})\smallsetminus 
\partial \Delta (\mathcal{K})$. In order to check whether or 
not $U$ is a critical point, $\wwi{k}$ will be replaced 
in~(\ref{eq: crystal}) by $m$ non-zero parallel rows
$\uui{\ell} = \alpha_\ell \wwi{k}$, where $\sum_{\ell 
\in\is{m}}\alpha_\ell = 1$ and $\alpha_\ell > 0$, $\ell \in 
\is{m}$.
Since the new rows $\uui{\ell}$ are all non-zero and 
strictly positive multiples of $\wwi{k}$,
\begin{enumerate}
\item[(A)] $\theta_{\wwi{k},\wwi{j}} = 
\theta_{\uui{\ell},\wwi{j}}, 
\; j \in \is{k}$, and $\theta_{\uui{\ell},\uui{\ell'}} = 0$ 
for 
all $\ell,\ell' \in \is{n}$.
\item[(B)] $\|\uui{\ell}\| = \alpha_\ell \|\wwi{k}\|, \; 
\ell \in \is{m}$.
\item[(C)] $\theta_{\uui{\ell},\vvi{j}} = 
\theta_{\wwi{k},\vvi{j}}$, for all $\ell \in \is{m}$, $j \in 
\is{d}$.
\item[(D)] $\big(W - V\big)^\Sigma = \big(U - V\big)^\Sigma$ (since $\sum_{\ell \in\is{m}}\alpha_\ell = 1$).
\end{enumerate}
We have $m$ new expressions replacing the right hand side of~(\ref{eq: crystal}) in case $i = k$:
\begin{align*}
&\sum_{j\in \is{\bar{k}}} 
\left(\frac{\|\wwi{j}\|\sin(\theta_{\uui{\ell},\wwi{j}})}{\|\uui{\ell}\|}\uui{\ell}
 -\theta_{\uui{\ell},\wwi{j}}\wwi{j}\right)-   \\
&\sum_{j\in 
\is{\bar{k}}}\left(\frac{\sin(\theta_{\uui{\ell},\vvi{j}})}{\|\uui{\ell}\|}\uui{\ell}
 - \theta_{\uui{\ell},\vvi{j}}\vvi{j}\right) + \pi\big(U - 
V\big)^\Sigma,
\end{align*}
where by $\sum_{j\in \is{\bar{k}}}$, we mean the sum over 
$\wwi{j}$ terms, $j \in \is{k-1}$, and $\wwi{\ell'}$ terms, 
$\ell' 
\in \is{m}$.
If $\ell \in \is{m}$, it follows from (A,B) that
\begin{align*}
&\sum_{j\in \is{\bar{k}}} 
\left(\frac{\|\wwi{j}\|\sin(\theta_{\uui{\ell}, 
\wwi{j}})}{\|\uui{\ell}\|}\uui{\ell}
 -\theta_{\uui{\ell}, \wwi{j}}\wwi{j}\right) =  
\sum_{j\in \is{d}} 
\left(\frac{\|\wwi{j}\|\sin(\theta_{\wwi{k},\wwi{j}})}{\|\wwi{k}\|}\wwi{k}
 -\theta_{\wwi{\ell},\wwi{j}}\wwi{j}\right),
\end{align*}
and from (B,C) that
\begin{align*}
&\sum_{j\in 
\is{k}}\left(\frac{\sin(\theta_{\uui{\ell},\vvi{j}})}{\|\uui{\ell}\|}\uui{\ell}
 - \theta_{\uui{\ell}, \vvi{j}}\vvi{j}\right) =
\sum_{j\in \is{k}}\left(\frac{\sin(\theta_{\wwi{k}, 
\vvi{j}})}{\|\wwi{i}\|}\wwi{i} - 
\theta_{\wwi{k},\vvi{j}}\vvi{j}\right)
\end{align*}
Noting (D), it follows that $U$ satisfies the critical point equations for $\ell \in \is{m}$. Along the same lines, but now
using the convexity condition $\sum_{\ell 
\in\is{m}}\alpha_\ell = 1$,
we verify that $U$ satisfies the critical point equations for $i \in \is{k-1}$.  Hence $U$ is a critical point of $\mathcal{L}$. \qed


\renewcommand{\is}[1]{{\mathbf{#1}}}

\section{Supplementary material for 
\pref{sec:hes_spec0}}\label{sec:hes_spec_pf}

We give a brief review of the technique used in this work 
to compute the Hessian spectrum. The introduction follows 
\cite{arjevanifield2020hessian,arjevanifield2021analytic} verbatim 
and is provided here for completeness. 

Suppose $V \subset \real^m$  is a linear subspace, with  Euclidean inner 
product induced from $\real^m$, and $(V,G)$ is an orthogonal 
$G$-representation.
\begin{lemma}{\cite[Lemma 7, Setion 
B.1]{arjevanifield2020hessian}}\label{lem: isow}
	The representation $(V,G)$ may be written as an orthogonal direct sum 
	$\bigoplus_{i=1}^m 
	(\oplus_{j=1}^{p_i}V_{ij})$ where
	$V_{ij} \subset V$, $(V_{ij},G)$ is irreducible, and $(V_{ij},G)$ is 
	isomorphic to 
	$(V_{\ell k},G)$ iff $i = \ell$, and $j,k \in \ibr{p_i}$.
	The subspaces $\oplus_{j=1}^{p_i}V_{ij}$ are unique, $i\in\ibr{m}$.
\end{lemma}

If $p_i = 1$, for all $i \in m$, the orthogonal decomposition given by the 
lemma is unique, 
up to order; otherwise the decomposition is not unique. In spite of the lack of 
uniqueness of Lemma~\ref{lem: isow}, in some cases 
there may be 
\emph{natural} choices of
invariant subspace for the irreducible components. This is exactly the 
situation for the 
isotypic decomposition of $(M(k,k),G)$, $G = S_p \times S_{k-p}$. This 
naturality allows us to give 
natural 
constructions of the matrices $M_i$, $i \in \ibr{m}$, used for
determining the spectrum of $G$-maps $A: M(k,k)\arr M(k,k)$.

The isotypic decomposition for $(M(k,k),S_k)$ is $2 \mathfrak{t}+3 
\mathfrak{s} 
+ 
\mathfrak{x}+\mathfrak{y}$, $k \ge 4$ (see \pref{sec:hes_spec0}).
The subspace of $M(k,k)$ determined by $2 \mathfrak{t}$ is the set of all 
$k\times k$ 
matrices $\mathcal{T}= \{T_{a,b}\dd a,b\in\real\}$
where the diagonal entries of $T_{a,b}$ all equal $a$ and the off-diagonal 
entries all 
equal $b$. There are many ways to write $\mathcal{T}$
as an orthogonal direct sum. For example, $\mathcal{T} = T_{1,1}\real \oplus 
T_{\frac{2}{k},-\frac{1}{k(k-1)}}\real$. However, there
is only one natural way: $\mathcal{T} = T_{1,0}\real  \oplus T_{0,1}\real$. 
Define $\mathfrak{D}^k_1 =  T_{1,0}$, $\mathfrak{D}^k_2 = T_{0,1}$. 
If we take the \emph{standard} realization of $(\mathfrak{t},S_k)$ to be 
$(\real,S_k)$,
where $S_k$ acts trivially on $\real$, then we have natural $S_k$-maps 
$\alpha_1,\alpha_2: \real \arr M(k,k)$ defined by $\alpha_i(t) = 
t\mathfrak{D}^k_i$, $i = 
1,2$.
If $A:M(k,k)\arr M(k,k)$ is an $S_k$-map, then $A$ restricts to the $S_k$-map 
$A_{\mathfrak{t}}: \mathcal{T} \arr\mathcal{T}$ and $A_{\mathfrak{t}}$ 
uniquely 
determines 
a 
$2 \times 2$-matrix $[a_{ij}]$ by
$A_{\mathfrak{t}}(\mathfrak{D}^k_i) = a_{i1}\mathfrak{D}^k_1 + 
a_{i2}\mathfrak{D}^k_2$, 
$i 
= 1,2$. The eigenvalues (and multiplicities in this case)
of $A_{\mathfrak{t}}:\mathcal{T}\arr\mathcal{T}$ are the same as the 
eigenvalues of 
$[a_{ij}]$.
If we choose a different orthogonal decomposition of $\mathcal{T}$, we get a 
different 
$2\times 2$-matrix that is similar to $[a_{ij}]$ and so has the same 
eigenvalues. The computation of the rest of the eigenvalues follows similarly.

\subsection{Proof of \pref{thm:xy_evs}} \label{sec:xy_evs_pf}
In \pref{sec:typeIorII} algebraic relations between the FPS 
coefficients are shown to reveal important 
information on the structure of regular families of critical 
points. In this section we show how these relations can 
be further used to evaluate the $\mathfrak{x}$- and the 
$\mathfrak{y}$-eigenvalues. The method is illustrated for 
regular families with $\base = 4$, $k=d+1$ and 
isotropy $\Delta (S_{d-1}\times S_1)$.

Referring to notation and results given in 
\pref{sec:typeIorII}, any type I or type II family of 
$\Delta (S_{d-1}\times S_1)$-critical points must satisfy $c_0\in\{\pm1\}$ and 
$c_1 = e_0 = e_1 = e_2 = e_3 = f_0 = f_1 = f_2 = f_3 = 0$. Below 
we shall assume that $c_0=-1$. The assumption is not needed, and 
is only introduced for ease 
of presentation. For non-zero $g_0$ and $i_0$, the 
Puiseux series of the 
eigenvalue associated to the $\fx$-representation is given 
by:
\begin{align}\label{eq:xev}
\lambda_\fx &= 
- \frac{c_{2}^{2}}{2 \pi} - \frac{c_{2} \sqrt{g_{0}^{2}}}{2 \pi} - 
\frac{c_{2} \sqrt{p_0^{2}}}{2 \pi} + \frac{c_{4}}{2 \pi} 
+ \frac{e_{4}^{2}}{4 \pi} - 
\frac{1}{\pi} + \frac{1}{4} + \frac{p_2 \sgn\prn{p_0}}{2\pi} + 
\frac{g_{2} \sgn\prn{g_{0}}}{2 \pi }\\
&+ d^{\frac{1}{4}} \left(\frac{c_{3}}{2 \pi} + \frac{p_1 \sgn\prn{p_0}}{2 
\pi} + \frac{g_{1} \sgn\prn{g_{0}}}{2 \pi }\right) + d^{\frac{1}{2}} 
\left(\frac{c_{2}}{2 \pi} + \frac{\sqrt{g_{0}^{2}}}{2 \pi} + 
\frac{\sqrt{p_0^{2}}}{2 \pi}\right) + O(d^{-\frac{1}{4}}).\nonumber
\end{align}
Thus, the expression $\lambda_\fx$ depends on FPS coefficients not 
determined in \pref{sec:typeIorII}. We show that 
$\lambda_\fx$ can be evaluated 
nonetheless, independently of the unknown coefficients. 

With FPS coefficients as above, 
\begin{align}\label{eq:spec_grad}
[d^{0}]\partder{1} &= \frac{c_{4}}{2 \pi}  + \frac{e_{4}^{2}}{4 \pi} + 
\frac{e_{4}}{4} + \frac{g_{0}}{4} + \frac{p_0}{4} - \frac{1}{2 \pi} + 
\frac{p_2 \sgn\prn{p_0}}{2 \pi} + \frac{g_{2} \sgn\prn{g_{0}}}{2 
	\pi},\nonumber\\
[d^{\frac{1}{4}}]\partder{1} &=  \frac{c_{3}}{2 \pi} + \frac{p_1 
	\sgn\prn{p_0}}{2 \pi } + \frac{g_{1} \sgn\prn{g_{0}}}{2 \pi },\nonumber\\
[d^{\frac{1}{2}}]\partder{1} &= \frac{c_{2}}{2 \pi} + \frac{\sqrt{g_{0}^{2}}}{2 
	\pi} + \frac{\sqrt{p_0^{2}}}{2 \pi},\\
[d^{0}]\partder{2}&= \frac{e_{4}}{4} + \frac{g_{0}}{4} + \frac{p_0}{4}.\nonumber
\end{align}
Substituting into \pref{eq:xev} gives
\begin{align}\label{eqn: x_ev_grad}
\lambda_\fx &= 
\frac{1}{4} - \frac{1}{2 \pi} + [d^0]\partder{1} - 
c_2[d^{\frac{1}{2}}]\partder{1} - [d^0]\partder{2}+
[d^{\frac{1}{4}}]\partder{1} d^{\frac{1}{4}}+
[d^{\frac{1}{2}}]\partder{1}d^{\frac{1}{2}} + O(d^{-\frac{1}{4}}).
\end{align}
In particular, $\lambda_\fx$ can be expressed in terms of $[d^{0}]\partder{1}, 
[d^{\frac{1}{4}}]\partder{1}, [d^{\frac{1}{2}}]\partder{1}$ and 
$[d^{0}]\partder{2}$. Therefore, by continuity, \pref{eqn: x_ev_grad} also applies 
for $g_0=p_0=0$. Since gradient entries vanish at critical 
points, so do their Puiseux coefficients and so 
$[d^{0}]\partder{1}, [d^{\frac{1}{4}}]\partder{1}, 
[d^{\frac{1}{2}}]\partder{1}$ and $[d^{0}]\partder{2}$  vanish, 
giving 
$\lambda_\fx = \frac{1}{4} - \frac{1}{2 \pi} + 
O(d^{\frac{1}{4}})$. \pref{eqn: x_ev_grad} not only gives 
the exact value of the $\fx$-eigenvalue to 
$O(d^{\frac{1}{4}})$-order but also 
describes its sensitivity to variations in the FPS coefficients. 
For example, it is seen that varying $c_2$ accounts for 
perturbations of order $O(d^{\frac{1}{2}})$. The derivation of 
the eigenvalue associated to the $\fy$-representation follows 
along the same lines, giving
\begin{align}
\lambda_\fy &= \frac{1}{4} + \frac{1}{2 \pi} + [d^0]\partder{1} - 
c_2[d^{\frac{1}{2}}]\partder{1} - [d^0]\partder{2}+
[d^{\frac{1}{4}}]\partder{1} d^{\frac{1}{4}}+
[d^{\frac{1}{2}}]\partder{1}d^{\frac{1}{2}} + O(d^{-\frac{1}{4}}).
\end{align}
Similar relations exist between \emph{criticality} and 
\emph{loss}.

\end{document}